\documentclass[12pt]{article}
\usepackage{graphicx}
\usepackage{amsmath}
\usepackage{amssymb}
\usepackage{authblk}
\usepackage[left=1in,right=1in,top=1in,bottom=1in]{geometry}
\usepackage{setspace}
\usepackage{float}
\usepackage[super,sort&compress,comma]{natbib}
\usepackage[version=3]{mhchem} 
\usepackage{graphicx} 
\usepackage[font={footnotesize}]{caption} 
\usepackage[labelfont=bf]{caption}
\usepackage{amsmath, amssymb}
\usepackage{natbib}
\usepackage{booktabs}
\usepackage{array}
\usepackage{subcaption}
\usepackage{multirow}
\usepackage{soul}
\usepackage{float}
\usepackage{siunitx}
\usepackage{tabularx}
\usepackage{ulem}
\usepackage{hyperref} 
\usepackage{xcolor}

\begin{document}

\title
{\textbf{SynLlama: Generating Synthesizable Molecules and Their Analogs with Large Language Models}}

\author{Kunyang Sun$^{1}$, Dorian Bagni$^{1,\Delta}$, Joseph M. Cavanagh$^{1,\Delta}$, Yingze Wang$^{1, \Delta}$, Jacob M. Sawyer$^{4}$, Bo Zhou$^{5}$, Andrew Gritsevskiy$^{6}$, Oufan Zhang$^{1}$, Teresa Head-Gordon*$^{1-3}$}
\date{}
\maketitle
\begin{center}
\vspace{-10mm}
$^1$Kenneth S. Pitzer Theory Center and Department of Chemistry, $^2$Department of Bioengineering, $^3$Department of Chemical and Biomolecular Engineering, University of California, Berkeley, CA, 94720 USA\\

$^4$Department of Chemistry, University of Minnesota, 207 Pleasant Street SE, Minneapolis, MN 55455, USA \\

$^5$Department of Pharmaceutical Sciences, University of Illinois Chicago, 833 S Wood St, Chicago, IL 60612, USA \\

$^6$Contramont Research, San Francisco, CA, 94158 USA

$^{\Delta}$authors contributed equally

corresponding author: thg@berkeley.edu
\end{center}

\begin{abstract}
\noindent
Generative machine learning models for exploring chemical space have shown immense promise, but many molecules they generate are too difficult to synthesize, making them impractical for further investigation or development. In this work, we present a novel approach by fine-tuning Meta's Llama3 Large Language Models (LLMs) to create SynLlama, which generates full synthetic pathways made of commonly accessible building blocks and robust organic reaction templates. SynLlama explores a large synthesizable space using significantly less data, and offers strong performance in both forward and bottom-up synthesis planning compared to other state-of-the-art methods. We find that SynLlama, even without training on external building blocks, can effectively generalize to unseen yet purchasable building blocks, meaning that its reconstruction capabilities extend to a broader synthesizable chemical space than the training data. We also demonstrate the use of SynLlama in a pharmaceutical context for synthesis planning of analog molecules and hit expansion leads for proposed inhibitors of target proteins, offering medicinal chemists a valuable tool for discovery.

\end{abstract}

\newpage
\section{Introduction}
\vspace{-2mm}
Chemical space is enormous, built up via the exponential rise in functional group combinatorics that define an increasing diverse set of molecules. Traditional approaches that design synthetic pathways of unseen molecules under well-controlled laboratory conditions have been made possible by decades of research in synthetic chemistry, as well as mechanistic studies of key reaction steps and reaction classification\cite{corey_general_1967, ihlenfeldt_computerassisted_1996}. Using the wealth of data accumulated in libraries of chemical reactions, expert systems\cite{corey_computer-assisted_1985, Kowalik2012} have been developed that deploy this knowledge to construct multi-step pathways to specified end-products. Such methods have become a key tool for the bench chemist, as illustrated by the Chematica software and and its follow on commercial product now known as Synthia\cite{Kowalik2012}.

With recent advances in artificial intelligence and deep learning, generative models have begun to contribute to enumerating molecules at the stoichiometric scale. After training on databases containing various small molecules representations\cite{pdbbind_2004,bindingdb1, irwin_zinc_2012,  gaulton_chembl_2012, enamine_bb}, string-based 1D generative models and structure-aware 3D \textit{de novo} methods have paved the way for quick exploration of greater swaths of unseen chemical space\cite{chithrananda_chemberta_2020, li_molbert_2021, mol-gen-gpt0, eckmann_limo_2022, wang_cmolgpt_2023, mol-gen-bg2, mol-gen-bg3, mol-gen-bg4, denovo_Guan2023TargetDiff,iminer, denovo_li2023ls, denovo_Luo2021SBDD, denovo_pocket2mol, denovo_zhang2022novo, denovo_Qian2022AlphaDrug, denovo_schneuing2022structure, denovo_zhang2023resgen,cavanagh_smileyllama_2024}. However, even with their exceptional generative capabilities, these models still face one major challenge: their proposed \textit{de novo} molecules lack practical guarantees of synthesizability, which limits their utility in practice\cite{sumita_hunting_2018, zhavoronkov_deep_2019, gao_synthesizability_2020}. For generative approaches in drug and materials discovery to fulfill their potential, ensuring synthetic feasibility is essential to bridge the gap between \textit{in silico} molecule design and the realistic applicability of computationally generated molecules.

In recognition of this dissonance, efforts have been made to address the problem of poor synthesizability of \textit{de-novo}-generated molecules. One line of research focuses on integrating empirical or deep-learning scoring functions\cite{ertl_estimation_sa_score_2009, coley_scscore_2018, vorsilak_syba_2020, thakkar_retrosynthetic_2021,wang_deepsa_2023,kim_dfrscore_2024,neeser_fsscore_2024}, such as the synthetic accessibility (SA) score \cite{ertl_estimation_sa_score_2009} and DeepSA score\cite{wang_deepsa_2023}, into the objective functions of learning algorithms. However, optimizing only the synthesizability score can still lead to the generation of unsynthesizable molecules because the scoring functions rely on identifying common fragments or reactive centers in molecules\cite{bilodeau_generative_2022}. In addition, they often assign bad scores to complex yet synthesizable molecules that require multi-step synthetic pathways, causing generative models to miss viable candidates \cite{skoraczynski_critical_2023}. Others have proposed improving synthesizability by building molecules from common molecular fragments\cite{podda_deep_2020, yang2021hit, chen_deep_2021, seo_molecular_2023}, but they still don't guarantee synthesis as these methods do not explicitly consider the reaction pathways to build the molecular candidates. Another line of research integrates the explicit use of computer-assisted synthetic planning (CASP) software\cite{genheden_aizynthfinder_2020, tu_askcos_2025} into the optimization\cite{guo_directly_2025}. However, the computational overhead is significant and the quality of the optimized molecules can be variable.\cite{Shen2021}.

Alternatively, proposing synthesizable molecular candidates using commercially available building blocks and commonly known organic reaction templates\cite{coley_robotic_2019, genheden_aizynthfinder_2020} offers better synthetic tractability over simple molecule scoring. Importantly, this strategy is appealing to bench and medicinal chemists, since it offers actionable synthesis pathways for them to examine, refine, and execute. Some recent models in this direction apply rule-based synthesis and optimization on building blocks or entire synthetic pathways to generate novel molecules with desired chemical properties\cite{Kowalik2012,li_deep_2024, swanson_generative_2024, cretu2025synflownet, seo_generative_2024, wang_clickgen_2024}. Other models condition on input molecules to propose synthetic pathways using commercially available building blocks and well-validated reaction templates for either full construction of the target molecule or the generation of structurally similar analogs in a forward synthesis manner within the predefined chemical search space.\cite{gao_amortized_2022,luo_projecting_2024, gao_generative_2024} For example, SynNet\cite{gao_amortized_2022} constructs synthetic trees via Markov Decision Processes (MDPs) and uses multilayer perceptrons to choose the next action space. More recent models such as ChemProjector\cite{luo_projecting_2024} and Synformer\cite{gao_generative_2024} use transformers to decode for the next action space and have achieved good empirical performances for target and analog molecule reconstruction. 

A compelling alternative is the use of Large Language Models (LLMs) due to their foundational nature and adaptability to downstream tasks. \cite{bommasani2022opportunitiesrisksfoundationmodels} LLMs inherently possess extensive chemical knowledge, and recent advancements have focused on extracting and applying this knowledge for predictive and optimization tasks using natural language guidance\cite{boiko_autonomous_2023, m_bran_augmenting_2024, yu2024llasmol, ramos_review_2024}. Furthermore, after fine-tuning, LLM models can perform as good or better than chemical language models trained solely on chemical representations, all while requiring less data.\cite{cavanagh_smileyllama_2024} The efficiency and unexpected performance gains from fine-tuning LLMs thus motivates us to explore their potential in more complex tasks, such as synthesis planning, which could pave the way for new chemical discoveries.

Herein, we present SynLlama, an LLM-based tool built on the open-source Llama-3.1-8B and Llama-3.2-1B foundation models ~\cite{dubey_llama_2024} to deduce synthetic routes for target molecules or structurally related analogs. Specifically, the LLM component of SynLlama operates as a constrained retrosynthesis module that breaks input molecules into building blocks (BBs) via well-validated (RXN) sequences, and the reconstruction module searches commercially available BBs based on LLM predictions and builds up molecules within a diverse yet synthesizable chemical space. As an illustration of utility, SynLlama demonstrates competitive performance in key tasks for drug discovery, including synthesis planning for target and analog molecules of pharmaceutical interest and expansion around existing molecular drug hits and leads. Moreover, because of its generative nature, the LLM component of SynLlama has the added ability to explore commercially available building blocks beyond the predefined synthetic space introduced during training - an ability that previous models lack. By integrating molecular design with synthetic feasibility, SynLlama represents a step forward in bridging computational chemistry with synthetic chemistry, providing  chemists with actionable and experimentally accessible molecular candidates.
\vspace{-3mm}

\section{Methods}
\begin{figure}[H]
\centering
\includegraphics[width=0.95\textwidth]{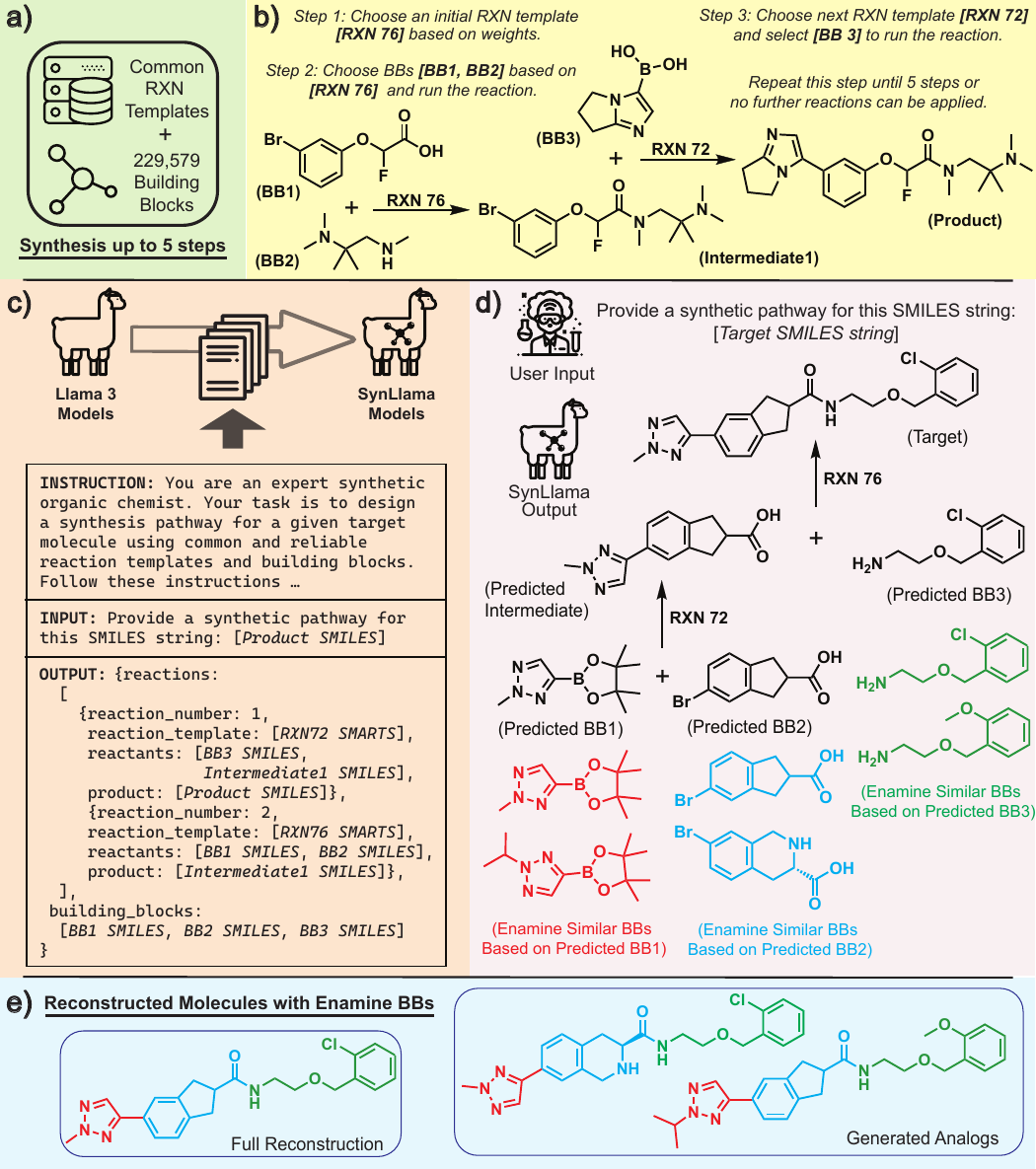}
\caption{\textbf{Overview of the SynLlama workflow including data generation, supervised fine-tuning, inference, and reconstruction.} (a). The predefined synthesizable chemical space of reaction templates (RXN) and building blocks (BBs) that covers billions of molecules. (b). An example synthesis data and its generation process from the defined synthesizable chemical space to create training examples. Here, RXN 76 represents amide coupling and RXN 72 represents Suzuki coupling. (c). A schematic representation of supervised fine-tuning that converts Llama 3 models to SynLlama models, along with the instruction, input, and output for the example synthesis in (b). (d). SynLlama's inference on an unseen test molecule. Black represents SynLlama's raw retrosynthetic output consisting of RXN sequences and predicted BBs, while colored BBs indicate the top two most similar BBs to the predicted ones from the Enamine building block library. Here, RXN 76 represents amide coupling and RXN 72 represents Suzuki coupling. (e). Reconstructed molecules using the predicted reaction sequences and similar building blocks from the Enamine building block library. In this example, all predicted building blocks are present in the Enamine library, allowing for the complete reconstruction of the input molecule and the generation of close analogs.}
\label{fig:synllama-wf}
\end{figure}

\vspace{-2mm}
The SynLlama workflow, illustrated in Figure \ref{fig:synllama-wf}, is designed to generate synthesizable compounds within an expanded chemical space. When an input molecule passes through this workflow, it can either be fully reconstructed through valid synthetic pathways, or the workflow will produce a structurally similar yet synthesizable analog along with its synthesis route. To transform general-purpose LLMs, like the Llama 3 models~\cite{dubey_llama_2024}, into expert models for synthetic pathways, we use three key components: 1) a reliable and diverse set of reaction data that covers a large synthesizable chemical space, 2) an efficient supervised fine-tuning (SFT) strategy to train a general-purpose LLM on these reaction data, and 3) a reconstruction algorithm that can convert the output of the fine-tuned LLM into valid synthesis routes, ensuring the proposed molecules lie within a commercially available chemical search space. These components are crucial for leveraging LLMs, which are known to perform well in diverse chemistry tasks~\cite{hendrycks2020mmlu, wang2024mmlupro}, to specialize in synthetic modeling. 

\subsection{Reaction Data for Training and Testing Sets}
\vspace{-2mm}
As illustrated in Figure \ref{fig:synllama-wf}(a), our defined chemical space for training consists of molecules that can be synthesized in at most five steps with Enamine building blocks\cite{enamine_bb} (BBs) and 2 sets of well-validated common organic reactions (RXNs). To define the training and testing BB data, we apply a time split whereby all Enamine BBs from the August 2024 release serve as the training BBs, and all new BBs from their February 2025 release that were not in the training set comprise the testing BBs. This procedure results in $\sim$230,000 BBs for training and $\sim$13,000 BBs for testing. Later in Section 3.1 we consider the target reconstruction for unseen molecules which are taken from the Enamine Diversity Set\cite{enamine_bb} and ChEMBL dataset\cite{gaulton_chembl_2012}.

We define two sets of reaction templates (RXN) which operate on the Enamine BBs. RXN 1 is formulated as a set of 91 reaction templates selected by Gao et al.\cite{gao_amortized_2022} from the works of Hartenfeller et al.\cite{hartenfeller_dogs_2012} and Button et al.\cite{button_automated_2019}. RXN 2 is comprised of 115 reactions selected by Gao et al.\cite{gao_generative_2024} that contains reactions used to create the Enamine REAL space\cite{enamine_bb} plus some reactions from RXN 1. All reaction templates are accessible to both sets of BBs, thus defining the training and testing chemical spaces.  As a result, there are $\sim$$10^{30}$ molecules within this space that can be represented by a synthesis path that comprises a sequence of BBs and RXNs.

To enumerate molecules within this space, we use an iterative approach by selecting RXN templates and searching for compatible BBs. Specifically, as demonstrated in Figure \ref{fig:synllama-wf}(b), the selection of the initial RXN is guided by a probabilistic model based on the number of compatible BBs. Within these compatible BB reactants, the initial BBs are selected at random to form an intermediate via the selected RXN template. This intermediate is then used to match for subsequent RXNs and recruits additional BBs to expand the molecular synthesis pathway until no further reactions are possible or the reaction reaches five steps. 

After training on these representations, the resulting LLM will be able to build powerful connections by mapping input molecules to a sequence of BBs and RXNs that creates linear synthesis routes for the testing sets of molecules. Hence, we also construct test sets for the more difficult case of branch  synthesis in which molecules have at least one or more reaction steps involving intermediates as reactants. To construct the branching synthesis test sets for both RXN 1 and 2  templates, we keep track of two synthesis trees at a time and check whether the intermediate molecules from both synthesis trees can react further with at least one reaction template from the corresponding RXN. We then filter for molecules that have at least one reaction step with reactants being intermediates to create the two test sets.
\vspace{-2mm}

\subsection{Supervised Fine-tuning and Inference from SynLlama}
\vspace{-2mm}
\label{subsec:sft}
To create the SynLlama model, we need to establish data generation protocols for supervised fine-tuning (SFT) of the Llama 3 models as schematically shown in Figure \ref{fig:synllama-wf}(c). When generating reaction data in text format, we choose to represent the BBs and intermediates along the synthetic pathway using SMILES\cite{SMILES} strings, while RXNs are explicitly defined in the SMARTS\cite{SMARTS} format. These structured chemical notations are designed to enhance SynLlama's ability to systematically identify and deconstruct bonds according to RXN templates, effectively dismantling input molecules into building-block-sized fragments. 

Since our goal is for SynLlama to learn to link molecules with their synthesis routes, our prompt-response pairs are structured according to retrosynthesis, as depicted in Figure \ref{fig:synllama-wf}(c) and shown in detail in Supplementary Figure S1. Such engineered prompts and responses allow the SynLlama model to learn to construct synthesis pathways for the input molecules by inferring sequences of BBs and RXNs, as well as the intermediate steps. While theoretically the model could predict BBs and RXNs without intermediates, we still include them in individual reaction steps in the hope of activating the inherent chemical knowledge in LLMs and enhancing their understanding of synthesis patterns. We have included some additional design choice analysis of forward synthesis versus retrosynthesis and whether to apply a drug-like product molecule filtering in the Supplementary Information.

We have considered both Llama-3.1-8B (8 Billion parameters) and Llama-3.2-1B (1 Billion parameters) for SFT using datasets of varying sizes. Specifically, Llama-3.1-8B is fine-tuned with datasets containing 100k and 500k synthesis routes, requiring 40 and 240 A-40 hours respectively. Llama-3.2-1B, on the other hand, is trained with datasets containing 500k and 2M synthesis entries, requiring approximately 60 and 240 hours respectively. Herein, we refer to the trained models as SynLlama-(parameter count)-(number of reactions trained) in the first part of Results. For example, SynLlama-1B-2M represents a model fine-tuned from Llama-3.2-1B with 2M synthesis routes. Further details of the SFT are provided in the Supplementary Information.

After training the SynLlama models, we apply the consistent prompt setup to perform inferences on molecules. For any given molecule, the SynLlama models predict reaction sequences in SMARTS format and generate SMILES strings for all the reactants, products, and BBs for the reactions they predict. During inference time, the instruction to SynLlama remains the same, and SMILES strings in the input section are substituted with ones specified by the user. As depicted in Supplementary Figure S1, the responses of the SynLlama models follow the output structures enforced by the prepared training prompt-response pairs. To be more specific, the output response section consists of two parts: reactions and building blocks. In the `reactions' component,  the model sequentially deconstructs the target molecule by breaking bonds using provided reaction templates in a retrosynthetic manner. At each step, it predicts a reaction template, along with the reactants and product of the reaction, continuing until no further reactions are possible. Then, in the `building blocks' section, the model compiles all building blocks, namely, reactants from each reaction that are not products in other reactions, identified from the `reaction' section. A visual representation of the inference process is illustrated in black ink in Figure \ref{fig:synllama-wf}(d). 
\vspace{-2mm}

\subsection{SynLlama Model Benchmarks}
\vspace{-2mm}
Since we are formulating the synthetic tasks using purely language-based modeling, where all reactions are expressed in SMARTS templates and molecules in SMILES strings, it is important to quantify the capacity of SynLlama for instruction following and comprehension of reaction chemistry. To assess SynLlama's ability to follow instructions, we select three benchmarking criteria as shown in Table \ref{tab:llm-benchmark}. The first is ``Valid JSON," which examines whether the output format is a parsable JSON following the fine-tuned templates that will be necessary for the downstream reconstruction algorithm. The second criterion is ``Template Memorization," which assesses the model's ability to memorize the provided reaction templates that define our synthesizable chemical space. Lastly, we benchmark on ``BB Selection," which evaluates whether the ``building blocks" section in the responses can accurately identify and select all the building blocks from the ``reactions" section of the responses. 

To assess SynLlama's comprehension of reaction chemistry, we focus on individual reactions and summarize the three critical aspects as: (1) the percentage of ``Valid SMILES" out of all SMILES strings in the responses, which is essential for assessing SynLlama's learning outcome of string-based chemical representations in general, (2) the percentage of ``Matched Reactants," which calculates whether the generated reactants match the reactant templates specified in the predicted reactions, and (3) the percentage of ``Good Products," which assess if the predicted product can indeed be generated by applying the proposed reaction templates onto the reactants. Overall, these six benchmarks can collectively assess SynLlama's capability to follow instructions and perform chemical reactions in string representations.

\begin{table}[H]
\small
    \centering
    \sisetup{detect-weight=true,mode=text}
    \begin{tabular}{@{}llcccc@{}}
        \toprule
        & & \multicolumn{4}{c}{Model Config} \\ 
        \cmidrule(lr){3-6}
        Dataset & Category & 8B-100k & 8B-500k & 1B-500k & 1B-2M \\ 
        \midrule
        \multirow{6}{*}{\begin{tabular}[c]{@{}c@{}}Training\\ Data\end{tabular}} 
        & Valid JSON & 96.20\% & 97.20\% & 96.60\%  & 98.00\% \\
        & Template Mem. & 99.95\% & 100.0\% & 100.0\% & 100.0\% \\
        & BB Selection & 99.80\% & 100.0\% & 99.72\% & 99.96\% \\ 
        \cmidrule(lr){2-6}
        & Valid SMILES & 94.74\% & 99.53\% & 95.17\% & 99.46\%\\
        & Matched Reactants & 78.62\% & 96.42\% & 80.19\% & 97.64\% \\ 
        & Good Products & 78.34\% & 96.58\% & 81.26\% & 98.58\%  \\ 
        \midrule
        \multirow{6}{*}{\begin{tabular}[c]{@{}c@{}}Testing\\ Data\end{tabular}} 
        & Valid JSON & 91.00\% & 94.20\% & 88.70\%  & 93.90\% \\
        & Template Mem. & 99.91\% & 100.0\% & 99.97\% & 100.0\% \\
        & BB Selection  & 99.75\% & 99.96\% & 99.73\% & 99.98\%\\ 
        \cmidrule(lr){2-6}
        & Valid SMILES & 94.37\% & 99.13\% & 87.66\% & 99.50\%\\
        & Matched Reactants & 77.51\% & 94.83\% & 65.63\% & 96.90\% \\ 
        & Good Products & 74.11\% & 94.16\% & 69.54\% & 96.39\%  \\ 
        \midrule
        \multirow{6}{*}{\begin{tabular}[c]{@{}c@{}}ChEMBL\\ Data\end{tabular}} 
        & Valid JSON & 98.80\% & 99.00\% & 99.20\%  & 99.00\% \\
        & Template Mem. & 99.90\% & 99.82\% & 99.37\% & 99.82\% \\
        & BB Selection & 99.57\% & 99.23\% & 99.50\% & 99.47\%\\ 
        \cmidrule(lr){2-6}
        & Valid SMILES & 92.02 \% & 96.38\% & 95.86\% & 95.23\%\\ 
        & Matched Reactants & 54.52\% & 69.25\% & 64.62\% & 70.93\% \\ 
        & Good Products & 67.69\% & 85.03\% & 75.81\% & 87.02\%  \\ 
        \bottomrule
    \end{tabular}
    \caption{\textbf{Benchmarks of SynLlama inferences on 1000 training, testing, and ChEMBL data.} We first select 1000 SMILES strings from the training examples, testing examples and the ChEMBL dataset, and then run inferences using SynLlama models trained on RXN 1. Benchmarking result for SynLlama models trained on RXN 2 can be found in Supplementary Table S1. The detailed descriptions of each benchmark can be found in the main text. Here, we run SynLlama inferences at $T = 0.1$ and $TopP=0.1$ to generate reproducible benchmarking results (see Supplementary Table S2).}
    \label{tab:llm-benchmark}
\end{table}

In Table \ref{tab:llm-benchmark}, all four trained SynLlama models are evaluated on both in-distribution training data and out-of-distribution testing and ChEMBL \cite{gaulton_chembl_2012} data to assess the benchmarks outlined above. In the instruction-following benchmarks, most models exhibit strong adherence (over 90 percent) to the fine-tuned response structure across all datasets. This impressive performance indicates that fine-tuning effectively retains the specified output structure when trained with over 100,000 samples. Furthermore, all four models successfully memorized the provided RXN templates and selected the building blocks (BBs) from all predicted reactants over 99 percent of the time. This capability further enhances the coupling effectiveness of the downstream reconstruction algorithm with the SynLlama raw output, as it only requires information about reaction sequences and predicted building blocks.

In the reaction chemistry benchmarking results, a clearer trend emerges: models, regardless of their parameter size, show improved comprehension of reaction chemistry in all three datasets as the amount of training data increases. Notably, most models maintain their performance from training to testing data, but exhibit a greater decline in ``Matched Reactants" and ``Good Products" performance when generalizing to the ChEMBL data. The reason behind this is that the testing data are generated in the same manner as the training data but with a different set of building blocks, while the ChEMBL data occupies a different chemical space, as previously noted by Gao et al. \cite{gao_amortized_2022}. Despite the reductions in their performance for ChEMBL molecules, as shown in Supplementary Figure S2, SynLlama-8B-500k and SynLlama-1B-2M can still generate complete and valid syntheses over 50\% of the time without any downstream processing. These results indicates that SynLlama’s raw results alone have potential utility for synthesis planning for unseen drug-like molecules.

When comparing SynLlama-8B-500k and SynLlama-1B-500k, we observe that the larger model demonstrates better performance when trained on the same amount of data. Although additional training data could further improve the 8B model based on the current trend, its higher computational cost makes this pursuit less practical. However, as the fine-tuning computational costs for SynLlama-8B-500k and SynLlama-1B-2M require approximately the same A40-GPU hours, and given the comparable benchmark performance between them, we decided to move forward with SynLlama-1B-2M, simplified as SynLlama, for the subsequent tasks due to its faster inference speed.
\vspace{-2mm}

\subsection{Reconstruction from Predicted Retrosynthesis}
\vspace{-2mm}
Using the predicted sequence of RXNs and BBs from SynLlama responses, we can synthesize the proposed target molecule or close analogs by applying the predicted reaction templates to the BBs in the inferred order, as shown in black ink in Figure \ref{fig:synllama-wf}(d). In some cases, the predicted BBs match known Enamine BBs, ensuring that the resulting molecules remain within an established chemical space for synthesis. However, due to SynLlama's generative nature, some predicted BBs are novel while still providing valid synthesis pathways, and we only report new BBs that can be purchased from other suppliers identified by Molport \cite{molport}. Therefore, while SynLlama primarily produces molecules within the predefined chemical space using Enamine BBs, its output also offers an alternative strategy for molecule construction. We will revisit this point in the Results section.

When the input molecule cannot be fully reconstructed, we generate analogs by mapping the predicted BBs from SynLlama to known Enamine BBs, thereby sampling molecules from the  well-defined Enamine chemical space. Under this scenario, we use nearest neighbor search algorithms with different molecular representations (SMILES and Morgan Fingerprints\cite{morgan1965unique}) to sample Enamine neighboring BBs from the predicted BBs, as illustrated in colored inks in Figure \ref{fig:synllama-wf}(d). Since in SynLlama's output, the RXN sequences are predicted concurrently with the BBs, our effective search space is constrained to Enamine BBs that can react through the specific RXN template. This smaller Enamine search space not only allows us to ensure the success rate of such forward syntheses but also allows us to effectively explore segments of the input molecule. Further details of the nearest neighbor search algorithms are provided in the Supplementary Information.

When constructing full synthetic pathways for reactions with multiple possible products, we select the product that most closely matches the predicted product based on SMILES string similarity. As shown in Figure \ref{fig:synllama-wf}(e), the reconstruction algorithm iteratively builds synthesis routes, utilizing all predicted BBs and RXN sequences to reconstruct or generate variations of the original molecule from the synthesizable chemical space. This reconstruction algorithm enables the SynLlama model to function as a generator for synthesizable molecules along with their corresponding synthetic pathways.
\vspace{-2mm}

\section{Results}
\vspace{-2mm}
We examine SynLlama's performance in synthesis planning of a diverse set of previously unseen compounds. We also explore the utility of the SynLlama workflow in real-world drug discovery applications, including its integration with generative algorithms to enhance the synthetic accessibility of proposed molecules while preserving their chemical properties and expanding the library of active compounds in the defined synthesizable space with as good or improved binding affinity metrics.
\vspace{-2mm}

\subsection{Synthesis Planning for Unseen Molecules}
\vspace{-2mm}
Having demonstrated that SynLlama models can reliably predict reaction sequences and building blocks in Table 1, we now use SynLlama to plan the synthesis of two groups of 1000 previously unseen molecules from the Enamine Diversity Set\cite{enamine_bb} and the publicly available ChEMBL database\cite{gaulton_chembl_2012}. These datasets are specific to drug-like molecules, unlike the training data used for SynLlama; the drug-related property distribution of both sets of molecules against the training data is shown in Figure S3. In this validation, we test whether known synthesizable molecules can be reconstructed accurately from the baseline and SynLlama models as summarized in Table 2.

We first consider the standard reconstruction approach used by algorithms such as SynNet\cite{gao_amortized_2022}, ChemProjector\cite{luo_projecting_2024}, and Synformer\cite{gao_generative_2024} to create target molecules or analogs using BBs exclusively from the Enamine library.  In this comparison, SynNet\cite{gao_amortized_2022} and ChemProjector\cite{luo_projecting_2024} serve as a baseline comparison for synthesizable chemical space coverage for RXN 1, whereas Synformer\cite{gao_generative_2024} is the baseline comparison on the expanded RXN 2 templates. As seen in the first column of Table \ref{tab:model-comparison} and Supplementary Table S3, when trained with their respective reaction sets and using only Enamine BBs, SynLlama outperforms all three methods while reducing the number of training data by 40-to 60-fold.

Although SynLlama yields higher percentages of successful ChEMBL reconstructions compared to SynNet and ChemProjector, and is on par with Synformer when only using Enamine BBs, there is a degradation of performance across all methods for ChEMBL compared to the Enamine Diversity set reconstructions. We attempted to improve upon the ChEMBL result by reformulating the training reaction data with extra filtering such that the product molecule distributions conform to a similar drug-like property distribution as the ChEMBL set as seen in Supplementary Figure S4. We then performed supervised fine-tuning following the same procedure as described in Section 2, but now with this filtered set of training data, in hope of better generating synthetic pathways for molecules with drug-like properties. However, as shown in Supplementary Table S3, there is now a performance loss over both the Enamine Diversity and ChEMBL drug-like targets. This result suggests that training on more diverse product molecules ultimately benefits synthesis planning for the drug-like targets more than specializing the LLM further with more restrictive training data. In addition, the curated ChEMBL data appears unique and outside the synthesizable chemical space made up of Enamine BBs and RXN templates.

\begin{table}[H]
\small
    \centering
    \renewcommand{\arraystretch}{1.05}  
    \setlength{\tabcolsep}{6pt}        
    \begin{tabular}{@{} ll ccc c @{}}
        \toprule
        \multirow{2.5}{*}{Dataset} & \multirow{2.5}{*}{Method} & \multicolumn{3}{c}{\# of Recon. Mol.} & \multirow{2.5}{*}{Morgan Sim.} \\
        \cmidrule(lr){3-5}
         & & Enamine BB & New BB & Total \\
        \midrule
        \multirow{5.5}{*}{\begin{tabular}[c]{@{}c@{}}Enamine\\ Diversity \\ Set\end{tabular}} 
            & SynNet & 110 & - & 110 & 0.57 \\
            & ChemProjector & 462 & - & 462 & 0.82 \\ 
            & Synformer & 660 & - & 660 & 0.91 \\ 
        \cmidrule(lr){2-6}
            & SynLlama(RXN 1) & 527 & 100 & 568 & 0.87 \\ 
            & SynLlama(RXN 2) & 691 & 232 & 741 & 0.92 \\
        \midrule
        \multirow{5.5}{*}{\begin{tabular}[c]{@{}c@{}}ChEMBL\\ Data\end{tabular}}  
            & SynNet   & 54  & - & 54 & 0.43 \\
            & ChemProjector & 133 & - & 133 & 0.60 \\ 
            & Synformer & 198 & - & 198 & 0.67 \\ 
        \cmidrule(lr){2-6}
            & SynLlama(RXN 1) & 165 & 95 & 223 & 0.66 \\
            & SynLlama(RXN 2) & 197 & 152 & 287 & 0.68 \\
        \bottomrule
    \end{tabular}
    \caption{\textbf{Comparison of synthesis planning performance among different methods.} Both the Enamine Diversity Set and ChEMBL Data are comprised of 1000 unseen molecules each. Details of each benchmark are described in the main text. The Morgan similarity scores include all analog molecules with successful synthesis pathways, as well as successfully reconstructed target molecules. The number of total training data and reaction set each method used for is: SynNet\cite{gao_amortized_2022} (\textbf{200K}, RXN 1) ChemProjector\cite{luo_projecting_2024} (\textbf{128M}, RXN 1), Synformer\cite{gao_generative_2024} (\textbf{85M}, RXN 2), and SynLlama (\textbf{2M}, RXN 1 \& 2). Further details are provided in Supplementary Tables S4 and S5, and identification of purchasable New BBs with Molport\cite{molport} are described in the Supplementary Information.}
    \label{tab:model-comparison}
\end{table}
\vspace{-3mm}

However, unlike baseline methods that only generate molecules using Enamine BBs, SynLlama has the extra capacity of reconstruct target molecules with commercially available BBs beyond Enamine due to its generative capabilities, even without specific training for this purpose.  As seen in Table \ref{tab:model-comparison} the `New BBs', restricted to those purchasable through Molport\cite{molport}, add possible synthetic pathways to reconstruct target molecules in all datasets and RXN templates. This also helps the ChEMBL data as well given that the molecules generated with the new BBs remain drug-like (Supplementary Figure S5). Since some target molecules can be synthesized through multiple pathways, either using only Enamine BBs or with the addition of New BBs, the `Total' column in Table \ref{tab:model-comparison} reflects the number of unique target molecules reconstructed with SynLlama. With these New BBs, SynLlama's best reconstruction rates increase to 74.1\% for the 1000 molecules in the Enamine Diversity Set, and encouragingly to 28.7\%  for the ChEMBL data. These results show that SynLlama learns reaction chemistry such that it can predict novel BBs to increase synthetic accessibility.

When the target molecule cannot be reconstructed, we assess the quality of the analog using a molecular similarity score between the target molecule and its most similar analog using Tanimoto similarity based on 4096-bit Morgan fingerprints\cite{morgan1965unique}. In Table \ref{tab:model-comparison}, the similarity metrics reported are average values of all generated molecules, including target molecules that are fully reconstructed (with a score of 1). Tables S6 and S7 also provide similarity metrics based on the 4096-bit Morgan fingerprints of Murcko scaffolds\cite{murcko_scaffold}, and Gobbi 2D pharmacophore fingerprints\cite{gobbi_1998}, while including or excluding target molecules that are fully reconstructed. Overall, these results collectively show that SynLlama is highly capable of planning synthesis for related analog molecules with very good similarity, aided most by increased synthetic pathways using purchasable building blocks.

Finally, Supplementary Table S3 also considers reconstruction performances for the Enamine Diversity Set and ChEMBL Data based on forward synthesis as opposed to retrosynthesis, and for test molecules derived from tree-like synthesis pathways. It is evident that SynLlama performs better for retrosynthesis relative to forward synthesis that is used more successfully by Synformer, and retrosynthesis also performs better than the baseline methods for branching synthetic pathways.

\subsection{Synthesizable Analog Search for \textit{De Novo} Molecules}
\vspace{-2mm}
Previous research has shown that molecules proposed by generative models for binding to protein targets often face challenges in both reliability and practical synthesizability\cite{sumita_hunting_2018, zhavoronkov_deep_2019, gao_synthesizability_2020}. In particular, medicinal chemists are often reluctant to devote time and expensive resources to specialized synthesis of hit molecules due to the high false positive rates arising from the unreliability of docking scores in drug discovery. Instead finding closely related compounds that are constrained to Enamine BBs allows for an inexpensive purchase to verify hits before further refinements are deployed to gain lead drug molecules. In this section, we demonstrate SynLlama's potential to bridge the gap between generative molecule design and practical synthesis for molecules that are optimized for drug-like applications such as binding assays.

Specifically, we consider two different generative methods for de novo drug molecules in Figure 2: a 1D-to-3D LSTM model, iMiner\cite{iminer}, which optimizes drug-likeness and AutoDock-vina\cite{Trott2010} docking scores, and Pocket2Mol\cite{denovo_pocket2mol}, a 3D graph transformer model which also optimizes AutoDock-vina\cite{Trott2010} docking scores. Using iMiner we generated 500 molecules for the SARS-CoV-2 Main Protease (SARS2 MPro)\cite{zhang2021potent}, and used Pocket2Mol to generate 500 molecules each for the protein targets Thrombin\cite{thrombin} and TYK2\cite{tyk2,tyk2_2}.  The first interesting observation is that only 1\% of generated molecules from both iMiner and Pocket2Mol can be synthetically reconstructed using SynLlama or ChemProjector, emphasizing that the generative models create difficult synthesis targets. Hence all 500 compounds for each protein target were then processed through SynLlama trained on RXN 2 to generate synthesizable analogs constrained to Enamine BBs. For all generated analogs we performed molecular docking with AutoDock-vina via using the same protocol as the original binders.

\begin{figure}[H]
    \centering
    \includegraphics[width=0.95\linewidth]{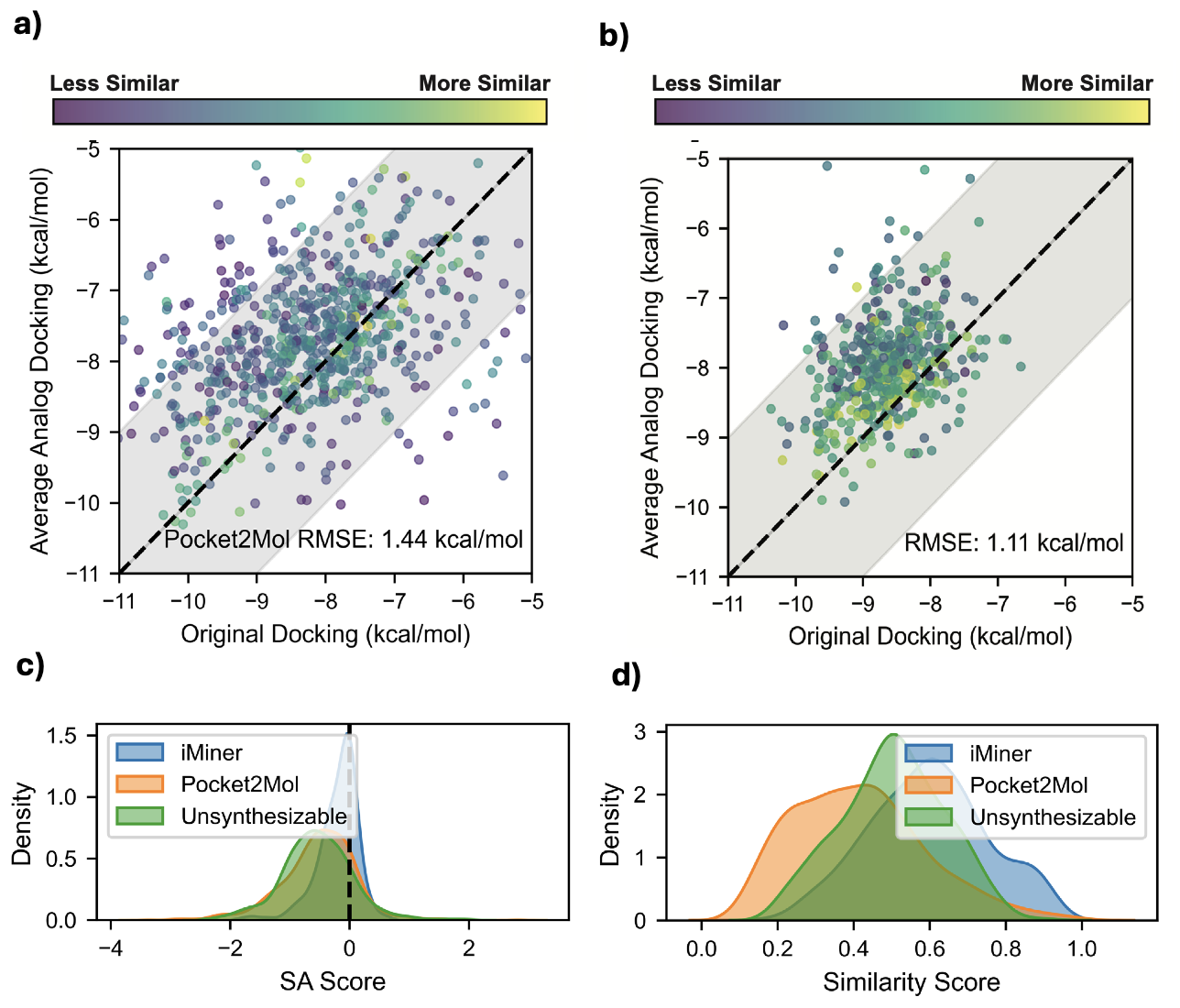}
    \vspace{-3mm}
    \caption{\textbf{SynLlama performance on generating synthesizable analogs for Pocket2Mol and iMiner proposed binders of SARS2 MPro\cite{zhang2021potent}, Thrombin\cite{thrombin}, and TYK2\cite{tyk2,tyk2_2}.} Correlation plot comparing docking scores of (a) Pocket2Mol and (b) iMiner generated molecules and the average Vina docking scores of ten most similar analogs from SynLlama trained with RXN 2. Each data point is color-coded by the average Morgan fingerprint similarity computed between the generated and analog molecules. The shaded area represents an energy uncertainty range of $\pm2 kcal/mol$ for docking\cite{Trott2010}. (c) Synthetic accessibility (SA) score distribution of Pocket2Mol, iMiner, and unsynthesizable molecules and SynLlama-proposed analogs. iMiner analogs generated with SynLlama trained on RXN 1 showed similar results as reported in Supplementary Figure S6. The kernel density in Supplementary Figure S7 further confirms our finding that the analogs consistently shift toward better SA without undermining the overall docking score distribution. (d) average Morgan fingerprint similarity score between the target molecules and their top-10 proposed analogs. }
    \label{fig:iminer_comparison}
\end{figure}

Figure \ref{fig:iminer_comparison}(a,b) shows the RMSE of the docking scores between the target compounds and the generated analogs are 1.44 kcal/mol and 1.11 kcal/mol for Pocket2Mol and iMiner, respectively, both of which are within the acceptable range of inherent docking score errors reported by Trott et al\cite{Trott2010}. Furthermore, as demonstrated in Figures \ref{fig:iminer_comparison}(c,d), the SA score distribution of the SynLlama analogs generated for thrombin and TYK2 from Pocket2Mol show a notable improvement in synthetic accessibility at the expense of reduced similarity below the benchmarks in Table 2. SynLlama slightly improves SA for iMiner generated compounds for SARS2 MPro while maintaining a good similarity score on par with the benchmarks in Table 2. The fact that iMiner uses a drug-likeness score as part of its loss function\cite{iminer} may explain its better performance, or perhaps SARS2 MPro is an easier protein target than Pocket2Mol's thrombin and TYK2 protein targets.

\vspace{-3mm}

\begin{figure}[H]
    \centering
    \includegraphics[width=0.9\linewidth]{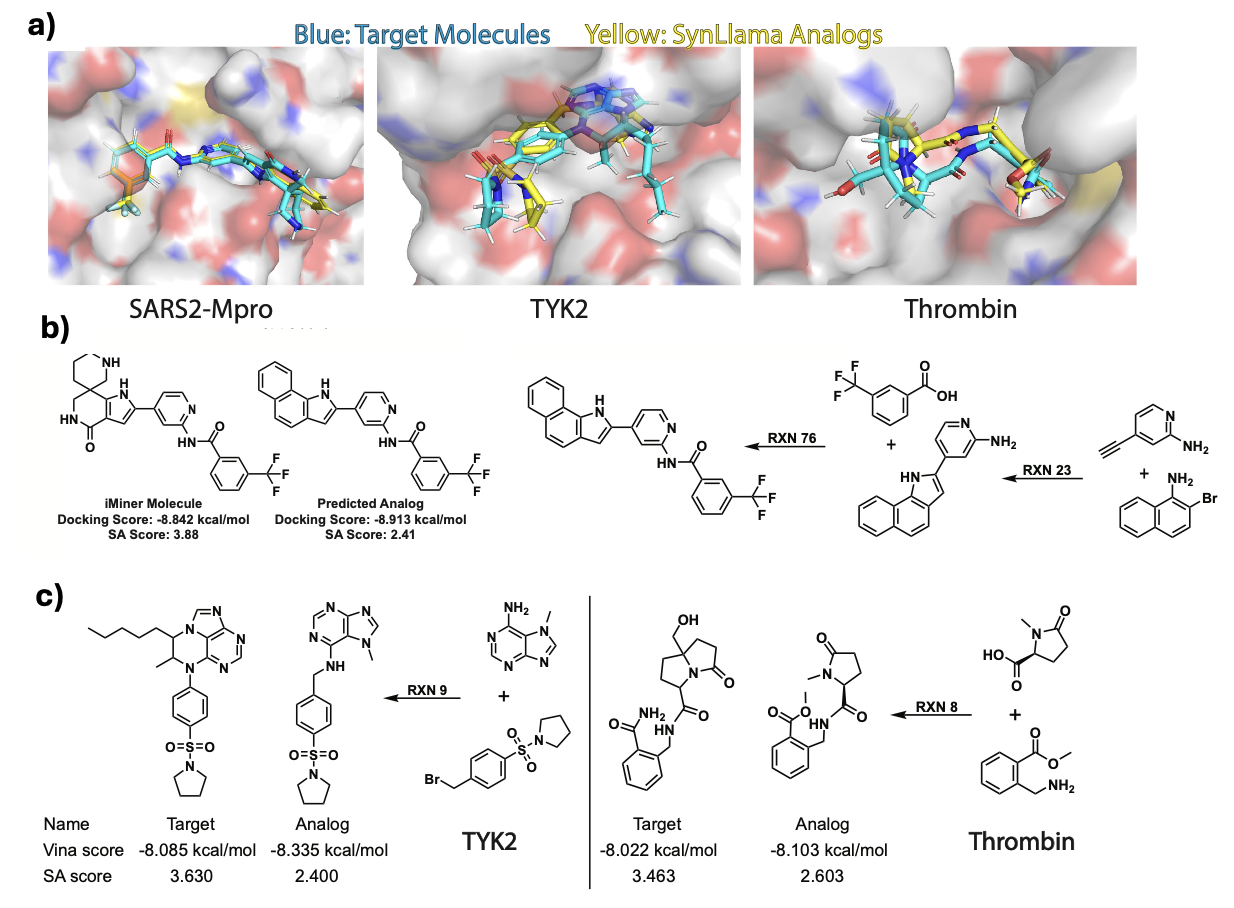}
    \vspace{-2mm}
    \caption{\textbf{Examples of synthesizable analog generation for SARS2 MPro using iMiner and TYK2 and Thrombin with Pocket2Mol}. (a) Docked pose visualization for all three protein targets. (b) Docking and SA scores for iMiner target and SynLlama analog for SARS2 MPro along with the predicted synthetic pathway. (c) Docking and SA scores for the Pocket2Mol targets and the SynLlama analogs for TYK2 and Thrombin along with their predicted synthetic pathways.}
    \label{fig:iminer_pairs}
\end{figure}

The better synthesizable analogs have good retention of the binding mode of the original generated molecules, visually confirmed by the representative docking poses for target-analog molecular pairs shown in Figure \ref{fig:iminer_pairs}(a) for the three proteins. Figure \ref{fig:iminer_pairs}(b,c) provides a few examples of the synthesis pathways for the analogs of the molecules derived from the generative molecules.
\vspace{-2mm}

\vspace{-2mm}
The final comparison consists of molecules that were identified as unsynthesizable by Gao et al\cite{gao_synthesizability_2020} via ASKCOS\cite{coley_robotic_2019}. As also seen in Figure 2, the similarity scores of SynLlama analogs for this set are better than that observed for the Pocket2Mol generative model, and the generated analogs show a significant decrease in SA score, representing SynLlamas effective strategy of improving synthetic accessibility via analog generation. In Supplementary Figure S8, we also highlight the few concrete examples of target-analog pairs where objective scores are maintained while target synthetic accessibility scores are substantially reduced. Overall, these collective results highlight SynLlama's utility in effectively proposing synthesizable analogs for \textit{de novo} molecules, thereby enhancing their synthetic accessibility without compromising their desired drug-related properties.


\subsection{Local Hit Expansion for Binder Molecules}
\vspace{-2mm}
Because SynLlama breaks down the original target molecule for synthesis into building blocks, by nature this method allows diverse exploration around parts of the molecular 

\vspace{-4mm}
\begin{figure}[H]
    \centering
    \includegraphics[width=0.9\linewidth]{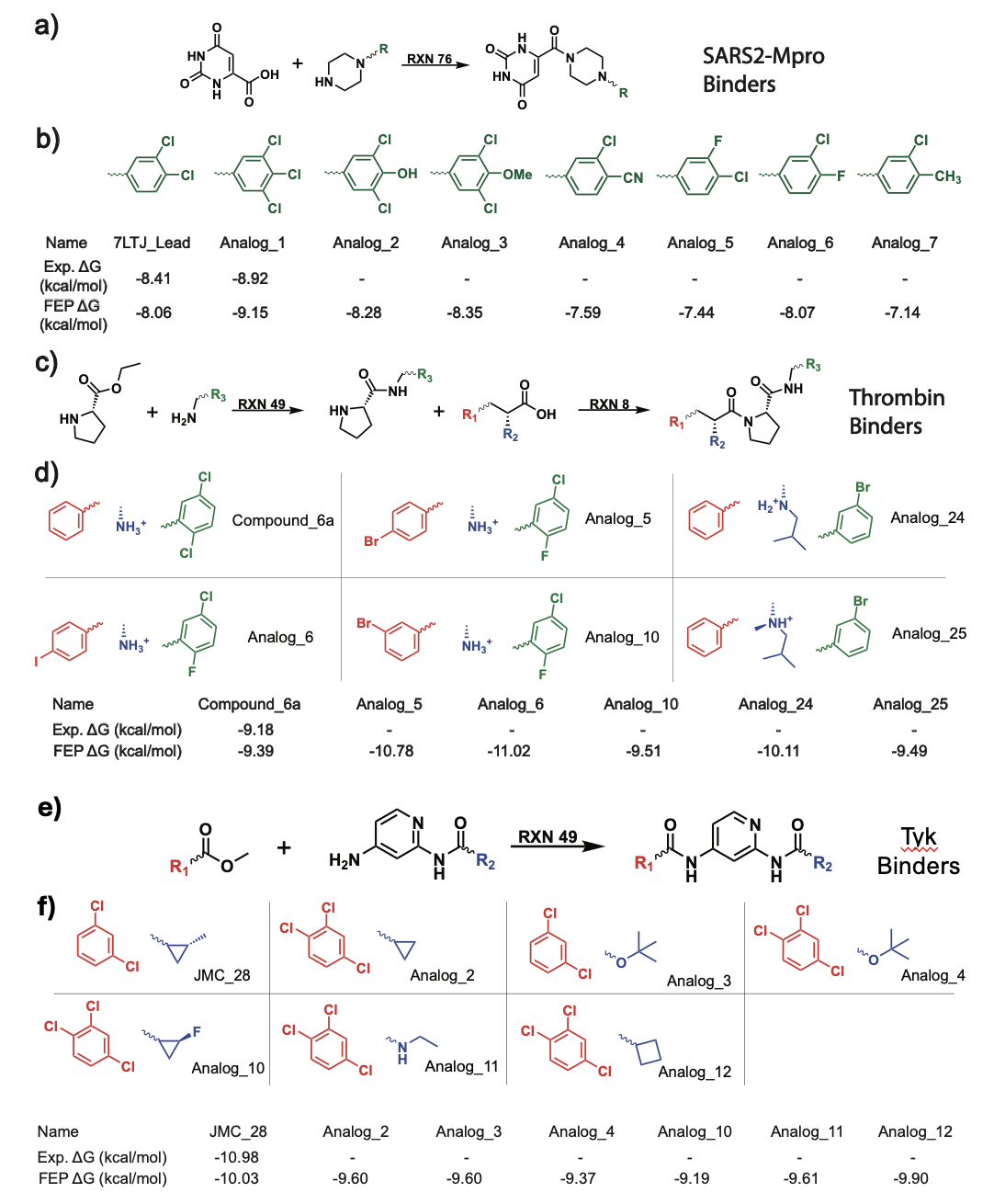}
    \vspace{-3mm}
    \caption{\textbf{Hit expansion of binders to SARS2 Mpro, Thrombin, and Tyk2 with SynLlama.} (a,c,e) Synllama-predicted synthetic pathways that expand on the hit molecules for each protein target. The places of substitution are labeled as R groups. (b,d,f) Binding free energies of the hit compounds and SynLlama-expanded analogs. Color scheme on the proposed substitution is the same as the predicted synthetic pathways. All potential binders either have a better FEP binding free energy or are within the 1 kcal/mol uncertainty range compared to the original hits.}
    \label{fig:fep}
\end{figure}
\vspace{-3mm}

\noindent
scaffold rather than only on a whole target molecule. In a final task, we apply SynLlama to expand on hit molecules for three protein targets used in the previous section, SARS2 Mpro\cite{7ltj}, Thrombin\cite{thrombin}, and TYK2\cite{tyk2,tyk2_2} to discover synthesizable molecules that have better relative binding free energies (RBFEs) confirmed by both experiments and accurate free energy perturbation (FEP) calculations.

As shown in Figure \ref{fig:fep}(a), the hit molecule for SARS2 Mpro (7LTJ\_Lead) has a core scaffold of uracil and ortho-dichlorobenzene connected by a piperizine linker. Inspired by an experimental hit expansion campaign by Kneller et al.~\cite{kneller_hit_expansion_7ltj}, we follow their practice to propose only functional group substitutions on the benzene ring while keeping the linker and uracil intact. In Figures \ref{fig:fep}(c) and \ref{fig:fep}(e), we use the best-performing molecules from the Schrodinger FEP benchmarking set\cite{schrodinger_bnechmark_github} as our hits for Thrombin and TYK2. To expand on these hit molecules, we first identify the maximum common scaffolds among each group of molecules in the FEP benchmarking set and use the identified scaffolds to guide our selection of analog molecules. Specifically, we use SynLlama to generate 50 synthesizable analogs constrained to only Enamine BBs of the hit compound and filter for molecules that retain the scaffold. In the end, we harvest a total of 8, 14, and 11 analog molecules that fulfill the criteria for SARS2 Mpro, Thrombin, and TYK2, respectively. The analogs are then placed in a pose configuration similar to the original hit molecule for downstream FEP calculations.

To verify the FEP results, we first choose $\sim$ 10 synthesized and experimentally tested molecules to benchmark the accuracy of FEP for all three systems. In Supplementary Figure S9, all the calculated FEP values show a good correlation with the experimental $\Delta G$ converted from IC50, with an average RMSE of less than $1\ \mathrm{kcal/mol}$. After this validation, we run FEP for the all proposed molecules to assess their binding affinities. As shown in Figures \ref{fig:fep}(b), \ref{fig:fep}(d), and \ref{fig:fep}(f), a significant portion of the proposed analogs (7 of 8, 5 of 14, and 6 of 11, respectively) showed potency compared to their parent hits for SARS2 Mpro, Thrombin, and TYK2. These results successfully demonstrate that SynLlama can propose diverse yet potent analogs when constrained on a molecular scaffold.

Moreover, the fact that all suggested analogs also come with predicted pathways using common reaction templates and purchasable BBs from Enamine suggests SynLlama's practical use for drug discovery. Because of this composite capacity of optimizing both potency and synthetic accessibility for small molecules, SynLlama successfully rediscovered Analog\_1, the most potent compound reported by Kneller et al. in their hit expansion campaign for SARS2 Mpro\cite{kneller_hit_expansion_7ltj}. Furthermore, for the Thrombin case, SynLlama explores the R2 substitution site, a region previously unaddressed by earlier molecular series, and demonstrates the generation of more potent molecules via FEP. These results confirm that SynLlama effectively explores local chemistry with readily available building blocks, providing a direct and efficient path for medicinal chemists to accelerate hit expansion.

\section{Discussion and Conclusions}
\vspace{-2mm}
Motivated by recent advances in LLMs for chemistry\cite{yu2024llasmol, boiko_autonomous_2023, cavanagh_smileyllama_2024, m_bran_augmenting_2024}, we aim to leverage data-efficient supervised fine-tuning (SFT) to transform the general-purpose Meta Llama 3 into SynLlama, an LLM-based generator capable of proposing synthesizable molecules and deducing synthetic routes for target molecules or their close analogs. Throughout the study, we successfully show that SynLlama can effectively explore a custom-defined chemical search space composed of around 230,000 Enamine building blocks (BBs) and well-validated organic reactions (RXNs), after it has been fine-tuned on synthetic pathway data sampled from this specified chemical space. What's more, despite utilizing nearly two orders of magnitude fewer synthetic pathways in training, SynLlama exhibits strong performance in key drug discovery tasks compared to existing models. Specifically, we have demonstrated that SynLlama can effectively aid in various stages of drug discovery that include synthesis planning, synthesizable analog generation for \textit{de novo} molecules, and local hit expansion.

Because SynLlama is built on a general-purpose LLM instead of training from scratch\cite{gao_amortized_2022, luo_projecting_2024, gao_generative_2024}, it offers a number of unique advantages and possibilities for further improvement. For example, when generating our fine-tuning data, we sampled from the predefined chemical search space of 230K Enamine building blocks (BBs) and two sets of reaction templates (RXNs), but we did not embed these extensive requirements in the context window of the LLM. As a result, for our largest dataset with only 2 million synthetic pathways, the model only saw each BB a few dozen times, while each RXN template appeared hundreds of thousands of times. Consequently, while SynLlama efficiently memorizes the allowed RXNs, it only captures the distribution of Enamine BBs, which enables SynLlama to extrapolate to unseen yet purchasable building blocks outside of Enamine. This generative ability surpasses other existing methods such as ChemProjector and Synformer that can only explore a predefined building block search space, such as within the Enamine Diversity Set\cite{enamine_bb}, and limits their ability to propose alternative synthesis pathway with novel building blocks.

In addition to its ability to extrapolate outside the training chemical space, the underlying Llama-3.2-1B used by SynLlama is relatively small and more predictive power would be expected if we train on larger LLMs with more data and compute power. However, we observe that a smaller LLM with fewer parameters can be turned into an expert model for complex tasks after SFT with relatively little data. This opens up opportunities to employ smaller expert models for various chemical tasks, benefiting from faster inference speeds, which can make these models more desirable. Moreover, optimal hyperparameters like temperature and top-p can vary between training and inference phases, depending on the downstream tasks. During inference, most valid raw outputs are generated under relatively low temperature and top-p settings. However, when the model is paired with reconstruction algorithms that require less strict adherence to reaction chemistry, higher temperature and top-p values can be used. This allows for a broader exploration of the Enamine chemical space, enabling the generation of more diverse and relevant analogs. This property is especially desirable in tasks that require extensive exploration, such as the hit expansion example we demonstrate. Another exciting direction is coupling SynLlama with another generative model, in which we have shown generates analogs while maintaining good docking scores and simultaneously shifting to better synthetic accessibility scores. This result suggests that SynLlama can serve effectively as a post-processor for other \textit{de novo} generative models, ensuring the production of more synthesizable compounds with clear reaction pathways.

Among the numerous opportunities that LLMs bring to the field of drug discovery, their natural language capabilities and recent advancements in reasoning are the most exciting features that allow users without coding expertise to interact directly with the models, effectively bridging the gap between computational methods and experimental research. We envision that expert users can employ prompt engineering and fine-tuning data to incorporate more realistic factors than those explored here. For instance, medicinal chemists could fine-tune LLMs within this generalizable SFT framework with building blocks and reaction templates of their own choice. In addition, they can consider synthesis cost, reaction conditions, improved selectivity, and protection factors at specific reaction steps for more detailed and powerful synthesis planning. We see our work as an initial attempt to demonstrate the effectiveness of LLMs in real experimental research, encouraging further studies for better utilization of these models.

\section{DATA AND CODE AVAILABILITY}
All the codes and data for SynLlama workflow are provided in a public accessible GitHub repository: https://github.com/THGLab/SynLlama under MIT License. 

\section{AUTHOR CONTRIBUTIONS}
K.S., D.B., J.C., and T.H.-G. conceived the scientific direction for SynLlama and wrote the manuscript. K.S. wrote the codes and trained the models. K.S., D.B., Y.W., and J.S. contributed to the result section. All authors provided comments on the results and manuscript.

\section{ACKNOWLEDGEMENT}
We thank Wenhao Gao for providing benchmarking data for SynNet and Synformer. We also express our gratitude to Shitong Luo for open-sourcing ChemProjector, whose GitHub repository served as a foundation for the development of the SynLlama GitHub. This work was supported by National Institute of Allergy and Infectious Disease grant U19-AI171954. This research used computational resources of the National Energy Research Scientific Computing, a DOE Office of Science User Facility supported by the Office of Science of the U.S. Department of Energy under Contract No. DE-AC02-05CH11231. 


\section{SUPPLEMENTARY INFORMATION}
Additional methodology and results are provided in the Supplementary Information, including details of SFT protocols; generation of training reaction data, hyperparameters used during inferences; baseline benchmarking procedures and results; procedures to check the purchasability and overall drug-related property distribution of novel building blocks; procedures to generate \textit{de novo} molecules via iMiner and Pocket2Mol; details to calculate various rewards functions (docking scores, SA scores, and FEP energies); results of additional benchmarking on reconstruction, LLM reliability, analog similarities; and the effect of different hyperparameters on LLM inferences and downstream Enamine reconstructions.

\bibliography{reference}
\bibliographystyle{naturemag}

\clearpage

\begin{figure}[p] 
\centering
\vspace*{\fill} 
\textbf{TOC Graphic}
\vspace{2em} 

\includegraphics[width=0.8\textwidth]{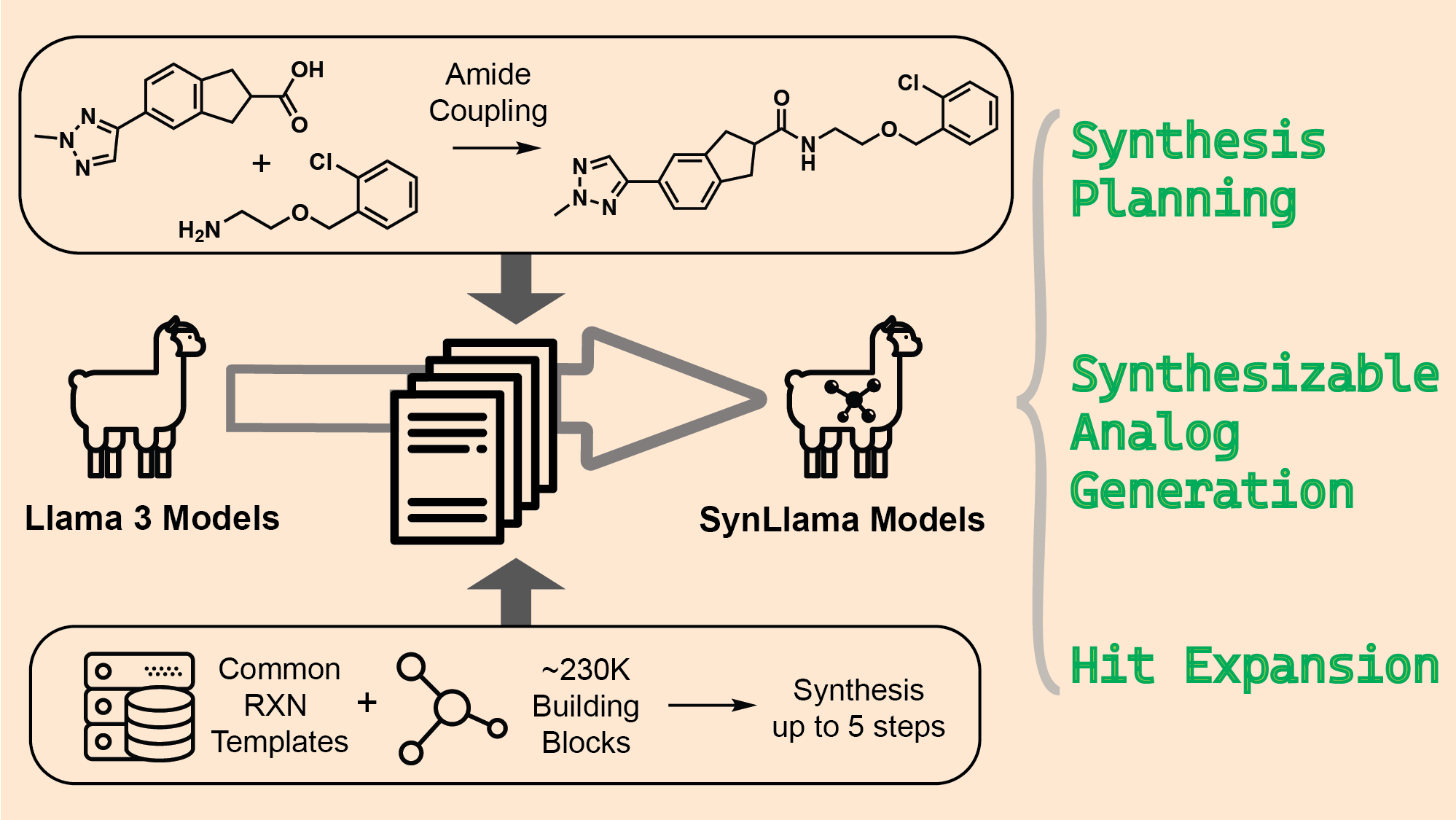} 

\vspace*{2em} 
\textbf{Synopsis: Fine-tuning on synthetic reactions from commercial building blocks and high-fidelity reactions creates a versatile LLM, SynLlama, for key drug discovery tasks.}
\end{figure}

\clearpage

\end{document}


\title
{\textbf{Supplementary Information\\
SynLlama: Generating Synthesizable Molecules and Their Analogs with Large Language Models}}

\author{Kunyang Sun$^{1}$, Dorian Bagni$^{1,\Delta}$, Joseph M. Cavanagh$^{1,\Delta}$, Yingze Wang$^{1, \Delta}$, Jacob M. Sawyer$^{4}$, Bo Zhou$^{5}$, Andrew Gritsevskiy$^{6}$, Oufan Zhang$^{1}$, Teresa Head-Gordon*$^{1-3}$}
\date{}
\maketitle
\begin{center}
\vspace{-10mm}
$^1$Kenneth S. Pitzer Theory Center and Department of Chemistry, $^2$Department of Bioengineering, $^3$Department of Chemical and Biomolecular Engineering, University of California, Berkeley, CA, 94720 USA\\

$^4$Department of Chemistry, University of Minnesota, 207 Pleasant Street SE, Minneapolis, MN 55455, USA $^5$Contramont Research, San Francisco, CA, 94158 USA\\

$^5$Department of Pharmaceutical Sciences, University of Illinois Chicago, 833 S Wood St, Chicago, IL 60612, USA

$^6$Contramont Research, San Francisco, CA, 94158 USA

$^{\Delta}$authors contributed equally

corresponding author: thg@berkeley.edu
\end{center}

\section*{Additional Methodology Details}

\textbf{Supervised Fine Tuning protocol.} After preparing the reaction data and prompt-response pairs from the training chemical space, we fine-tune Llama-3.1-8B (8 Billion parameters) and Llama-3.2-1B (1 Billion parameters) using the Axolotl package~\cite{dubey_llama_2024,noauthor_axolotl-ai-cloudaxolotl_nodate} for 1 epoch. LLMs with more parameters require more resources to train and use, but they also typically perform better on a variety of tasks, which we consider in Results. For our SFT approach, we apply Low-Rank Adaptation (LoRA) with a rank of $r=32$ and $\alpha=16$ to the linear layers of the model.\cite{hu_lora_2021} We use FlashAttention-2\cite{dao2023flashattention2}, with the Adam optimizer\cite{kingma_adam_2017}, cross-entropy loss, and a cosine learning rate scheduler with a maximum learning rate of $2 \times 10^{-5}$. 

\vspace{3mm}
\noindent
\textbf{Molecule Generation using Enamine BBs.} When searching for the nearest neighbors of BBs, a natural choice is to perform a string-level similarity search based on SMILES strings, as this is the native format of SynLlama responses. For each RXN template, we systematically process all SMILES strings of its compatible building blocks that can participate in the reaction. First, we extract the full vocabulary of SMILES tokens and generate an n-gram representation by considering all possible consecutive token pairs (bigrams) and triplets (trigrams). Next, we identify the 1024 most frequently occurring n-grams across SMILES strings of all compatible BBs to form a representative token set for each individual RXN template. To facilitate efficient retrieval, we structure search trees based on the term frequency-inverse document frequency (TF-IDF) scores \cite{sparckjones1972statistical} of these n-grams, prioritizing highly informative substructures and accelerating inference. Consequently, when a new SMILES string of the predicted BB is introduced, it can be efficiently processed through the tree, yielding a list of the top K matching SMILES strings.

In addition, Gao et al.\cite{gao_amortized_2022} investigated using Morgan fingerprints\cite{morgan1965unique}, a molecular representation capturing local chemical environment, to search for the nearest neighbors of BBs based on their Tanimoto similarity \cite{rogers_computer_1960}. Similarly to that stated above, for each RXN template, we also build a separate search tree for all compatible Enamine BBs using 256-bit Morgan fingerprint representation with a searching radius of 2. Our empirical observations indicate that combining the top K molecules from both the SMILES and Morgan fingerprint methods offers better performance than relying on the top 2K molecules from a single method. However, since we are working with an LLM model, the generated SMILES strings still have a small chance of being invalid, which prevents us from calculating their Morgan fingerprints. Therefore, we employ the both combined TF-IDF and Morgan fingerprint search trees when dealing with valid molecules, and revert to only a SMILES-based search when the generated SMILES strings are invalid.


\vspace{3mm}
\noindent
\textbf{LLM Inference Hyperparameters for Various Tasks.} A key advantage of SynLlama, and LLMs in general, is their sensitivity to variations in hyperparameters, such as temperature ($T$) and top-p ($TopP$), which can significantly impact the performance of reconstruction and analog similarity. As shown in Supplementary Figure S2, SynLlama's raw outputs exhibit enhanced reaction chemistry comprehension when inferences are run at lower $TopP$ and within a reasonable range of $T$ for both test sets. This configuration allows SynLlama to explore purchasable building blocks outside the Enamine library while maintaining synthesis validity. Conversely, increasing $T$ and $TopP$ generally reduces SynLlama's ability to generate valid syntheses in its raw outputs. However, as Supplementary Figure S2 also illustrates, inferring with higher $T$ and $TopP$ values than the optimal settings in raw outputs often leads to better overall average maximum similarity scores for reconstruction with Enamine BBs along. Nonetheless, excessively high settings can increase the failure rate. 

Based on empirical observations, we recommend specific combinations of $T$ and $TopP$ that effectively span a broad spectrum of tasks. These combinations optimize SynLlama's performance by balancing exploration and precision during inference.

\begin{itemize}
    \item \textbf{Frozen}: $T = 0.1, TopP = 0.1$, repeated once. This setting prioritizes deterministic generation, ensuring minimal variability and high reproducibility.
    \item \textbf{Low}: $T = 0.6, TopP = 0.5$, repeated multiple times. This configuration allows for limited exploration while maintaining a degree of precision.
    \item \textbf{Medium}: $T = 1.0, TopP = 0.7$, repeated multiple times. This setting balances exploration and diversity, generating outputs with moderate randomness.
    \item \textbf{High}: $T = 1.5, TopP = 0.9$, repeated multiple times. This configuration promotes high diversity and creativity in generation but may introduce more variability in results.
\end{itemize}

\noindent We define different sampling strategies based on these core settings:
\begin{itemize}
    \item \textbf{Frugal Sampling}: A total of 4 inferences.
    \begin{itemize}
        \item $T = 0.1, TopP = 0.1$, repeated one time.
        \item $T = 0.6, TopP = 0.5$, repeated one time.
        \item $T = 1.0, TopP = 0.7$, repeated one time.
        \item $T = 1.5, TopP = 0.9$, repeated one time.
    \end{itemize}
    \item \textbf{Greedy Sampling}: A total of 10 inferences.
    \begin{itemize}
        \item $T = 0.1, TopP = 0.1$, repeated one time.
        \item $T = 0.6, TopP = 0.5$, repeated two times.
        \item $T = 1.0, TopP = 0.7$, repeated three times.
        \item $T = 1.5, TopP = 0.9$, repeated four times.
    \end{itemize}
    \item \textbf{Frozen Only}: A total of 1 inference.
    \begin{itemize}
        \item $T = 0.1, TopP = 0.1$, repeated one time.
    \end{itemize}

    \item \textbf{Low Only}: A total of 5 inferences.
    \begin{itemize}
        \item $T = 0.6, TopP = 0.5$, repeated five times.
    \end{itemize}

    \item \textbf{Medium Only}: A total of 5 inferences.
    \begin{itemize}
        \item $T = 1.0, TopP = 0.7$, repeated five times.
    \end{itemize}

    \item \textbf{High Only}: A total of 5 inferences.
    \begin{itemize}
        \item $T = 1.5, TopP = 0.9$, repeated five times.
    \end{itemize}

\end{itemize}

\noindent
\noindent\textbf{Baseline Benchmarking Details.} Since the Enamine BB catalog constantly updates new BBs and does not store historical data, we cannot access the exact training BBs  used in training for the baseline methods ChemProjector\cite{luo_projecting_2024} and Synformer\cite{gao_generative_2024}. The only training set data we had access to is described in Section 2.1 in the main document, and is $\sim$ 3\% (10k) more compared to that available to Synformer and Chemprojector (cutoff at October 2023). However, we highlight that 97\% of our training data is identical to their previous work and the
newly added building blocks (which is equivalent to a time split) show a similar distribution as
the rest of the training BBs. We now have this comparison in Supplementary Figure 
S5. Therefore, we can fairly say that our training data are very similar to the baseline methods to which we compare. For a fair comparison at the inference stage, we provide ChemProjector and Synformer the same set of building blocks (cutoff at Feb. 2025) that SynLlama had access to during inference time. In Tables 2 and S3, we assess the performance of the trained baseline models with this more recent set of building blocks.\\

\noindent\textbf{Checking Commercial Availability of Building Blocks via Molport.} In Results, we used the Molport platform to check whether a predicted BB is commercially available or not. Initially, we compiled a list of building blocks for searching and used the `List Search' tab in the Molport website (https://www.molport.com/shop/swl-step-1) to check their availability. Once the SMILES strings were entered into the search interface, we set the search criteria to a minimum acceptable quantity of 500 mg and match types restricted to `Exact' and `Perfect' to search in the database of `screening compounds' and `building blocks.' Once the search completed, we downloaded the excel file under the `Selected Items' column from the List Search result tab (https://www.molport.com/shop/swl-requests), which contained both the commercially available compounds and information about the supplying vendors.

\vspace{3mm}
\noindent
\noindent\textbf{Calculation of SA Scores.} We calculate SA scores for both the iMiner-proposed molecules and SynLlama-generated analogs using the oracle functions named `SA' implemented in the TDC Commons package\cite{Huang2021tdc}. 

\vspace{3mm}
\noindent\textbf{iMiner-Generated Molecules and Docking Procedures for Analogs.} The iMiner algorithm\cite{iminer}, an 1D string-based LSTM model for SELFIES\cite{Krenn2020} string generation, was employed in this study. The molecules generated by iMiner are optimized for 3D shape complementarity using a composite objective function comprising the AutoDock Vina\cite{Trott2010} docking score, as well as a custom-defined druglikeness score\cite{iminer}. 

For molecular docking tasks, we obtained the SARS-CoV-2 Mpro crystal structure (PDB ID: 7L11\cite{zhang2021potent}) from the Protein Data Bank\cite{berman2000protein} and processed it with PDBFixer\cite{eastman2013openmm} to add missing hydrogens and remove heteroatoms. The docking grid was centered at the geometric center of the ligand (XF1) from the corresponding PDB file ([x = -22, y = -4, z = -28]) using a cubic box with 20 Å sides. Both proteins and ligands were converted to PDBQT format using Meeko (https://github.com/forlilab/meeko). Docking was performed with AutoDock Vina using an exhaustiveness parameter of 64, and the best pose for each ligand was recorded. This protocol was consistently applied during both iMiner training and analog docking assessments.

The custom drug‐likeness score is a composite score that evaluates 13 key molecular properties derived from the ChEMBL database. These properties capture both basic structural features and nuanced physicochemical characteristics, including the fraction of sp³-hybridized carbons, the total number of heavy atoms, and the fraction of non-carbon atoms within these heavy atoms. Additionally, the score accounts for the counts of hydrogen bond donors and acceptors, the number of rotatable bonds, and the balance between aliphatic and aromatic rings, along with molecular weight. Complementing these are parameters such as the approximate log partition coefficient (alogP), polarizable surface area (PSA), the number of structural alerts, and the size of the largest ring present in the molecule. Each property contributes to the overall score through a weight that is inversely proportional to the entropy of its distribution in the ChEMBL database: properties with narrower and more informative distributions exert a stronger influence. By summing the log likelihoods of these properties with their respective weights, the score effectively biases the generative model to produce molecules that closely mimic the drug-like profiles observed in established therapeutics, ensuring that the exploration of chemical space remains focused on compounds with favorable bio-availability and efficacy profiles.

\vspace{3mm}
\noindent
\textbf{Pocket2Mol Generation.} \textit{De novo} generation with Pocket2Mol was performed for Thrombin and TYK2 targets using codes from the Pocket2Mol github repository\cite{p2m_github}. Three default settings specified in \texttt{configs/sample\_for\_pdb.yml} were modified to generate at least 1000 molecules in one single run: \texttt{num\_samples:1000, beam\_size:500, max\_steps:100}. The protein structure files were downloaded from the Schrödinger FEP benchmark github repository\cite{schrodinger_bnechmark_github}. The pocket center was set to (-4.0, 26.5, -30.0) for TYK2 and (17.0, -12.5, 22.5) for Thrombin.

\vspace{3mm}
\noindent
\textbf{Unsynthesizable Molecules Identificaiton and Reward Calculation.} To access the list of the unsynthesizable molecules, we query the first 50 top-scoring molecules that were identified as unsynthesizable by ASKCOS\cite{tu_askcos_2025} for each property category listed in this csv (\url{https://github.com/wenhao-gao/askcos\_synthesizability/blob/master/results/goal\_hard\_cwo.csv}). There are a total of 10 different individual rewards, including 7 multi-property objectives (MPOs) centering around 7 different drug targets (Osimertinib, Fexofenadine, Ranolazine, Perindopril, Amlodipine, Sitagliptin, Zaleplon), Valsartan SMARTS, and 2 Hopping (Scaffold and deco). We used the TDC Commons package \cite{Huang2021tdc} to score both the original molecules and the generated analogs for their corresponding property category.

\vspace{3mm}
\noindent\textbf{Free Energy Perturbation (FEP) Protocols.} The relative binding free energies are calculated using GPU-accelerated AMBER22\cite{amber22} (\texttt{pmemd.cuda.MPI}). AMBER14SB\cite{amber14} and OpenFF-2.1.0\cite{OpenFF2} were used to parametrize the protein and the ligand, respectively. The SARS-CoV-2 Mpro protein structure (PDB code: 7LTJ) was downloaded from RCSB PDB and prepared with PDBFixer\cite{pdbfixer} to assign side-chain protonation states at pH=7.4 and add hydrogens. H163 was manually set to be its variant HIE (hydrogen added on N$\epsilon$) to ensure the correct hydrogen bonding with the ligand. For TYK2 and Thrombin, their protein structures were downloaded from the github repository of Schrödinger benchmark dataset\cite{schrodinger_bnechmark_github}. A sub-module \texttt{app.Modeller} in OpenMM\cite{openmm8} was used to immerse the protein-ligand complexes and unbound ligands  in a cubic water box with $15\mathrm{\mathring{A}}$ buffer size and add ions ($\mathrm{Na^+, Cl^-}$) to neutralize the system and maintain 0.15M ionic strength. For perturbations involving charge changes, the alchemical water method\cite{chen2018accurate} was used to eliminate the artifacts in PME simulation of system with net-charges.

We used 16 unevenly distributed lambdas (0.0, 0.174, 0.226, 0.265, 0.330, 0.383, 0.432, 0.477, 0.522, 0.568, 0.617, 0.670, 0.735, 0.774, 0.826, 1.0) to transform the initial state to the final state in the free energy. This lambda settings was designed to maximize the phase space overlap between adjacent states with the second-order smooth-step function introduced. The transformations were performed with the modified SSC(2) softcore potentials ($m = n = 2, \alpha_{\mathrm{LJ}} = 0.5,  \alpha_\mathrm{Coul} = 1$)\cite{ssc2_tsai2023amber}. Kartograf\cite{ries2024kartograf} algorithm was used to determine the common core region (SC) and soft core region (SC) atoms. 

Each lambda state was subjected to the following simulation protocol to equilibrate the system: (1) energy minimization without any constriants; (2) heating from 0 to 100 K at constant volume and temperature (NVT) ensemble over 20 ps, followed by MD at constant pressure and temperature (NPT) ensemble at 100 K for 20 ps; (3) heating to 200 K at NVT ensemble over 20 ps followed by another 20 ps at NPT ensemble at 200 K; (4) heating to 298.15 K at NVT ensemble over 20 ps followed by another 20 ps at NPT ensemble at 298.15 K; (5) another pre-prodcution equilibrium run at NPT ensemble for 500 ps. During the equilibration steps 2-4, restraints ($\mathrm{5\ kJ\cdot mol^{-1}\cdot \mathring{A}^2}$) were applied to heavy atoms on the solute. Finally, a 5-ns production run was performed for each lambda state with the ACES enhanced sampling method\cite{lee2023aces} and replica exchange was attempted every 0.5 ps. All the simulations employed 4 fs time step with the mass of solute hydrogens repartitioned to 3 amu\cite{hmr_hopkins2015long}. MBAR algorithm implemented in \texttt{alchemlyb}\cite{alchemlyb_Wu2024} was used to estimate the free energy change between two states and yield $\Delta\Delta G$. Then, the maximum likelihood estimation (MLE) method\cite{mle_xu2019optimal} was used to calculate the absolute binding free energy ($\Delta G$) of each ligand and the $\Delta G$ was shifted to make the average of calculated $\Delta G$ of the ligands equal to the average of their experimental $\Delta G$:

\begin{equation*}
    \sum_{i}\Delta G_\mathrm{pred}^{(i)} = \sum_{i}\Delta G_{\mathrm{expt}}^{(i)} = \sum_{i}RT\ln \mathrm{IC_{50}}^{(i)}
\end{equation*}

All the system preparation and analysis were performed with an in-house package named \texttt{easybfe} that automates the whole workflow and manage the calculations with high-performance computing platforms, and it will be described in a future publication in details.

\section*{Supporting Tables}

\begin{table}[H]
\small
    \centering
    \sisetup{detect-weight=true,mode=text}
    \begin{tabular}{@{}llcc@{}}
        \toprule
        Dataset & Category & SynLlama(RXN 1) & SynLlama(RXN 2) \\ 
        \midrule
        \multirow{6}{*}{\begin{tabular}[c]{@{}c@{}}Training\\ Data\end{tabular}} 
        & Valid JSON & 98.00\% & 98.20\%\\
        & Template Mem. & 100.0\% & 100.0\%\\
        & BB Selection & 99.96\% & 99.96\%\\ 
        \cmidrule(lr){2-4}
        & Valid SMILES & 99.46\% & 99.70\%\\
        & Matched Reactants & 97.64\% & 97.95\%\\ 
        & Good Products & 98.58\% & 97.97\%\\ 
        \midrule
        \multirow{6}{*}{\begin{tabular}[c]{@{}c@{}}Testing\\ Data\end{tabular}} 
        & Valid JSON & 93.90\% & 94.60\%\\
        & Template Mem. & 100.0\% & 100.0\%\\
        & BB Selection  & 99.66\% & 100.0\%\\ 
        \cmidrule(lr){2-4}
        & Valid SMILES & 99.50\% & 99.46\%\\
        & Matched Reactants & 96.90\% & 97.25\%\\ 
        & Good Products & 96.39\% & 96.19\%\\ 
        \midrule
        \multirow{6}{*}{\begin{tabular}[c]{@{}c@{}}ChEMBL\\ Data\end{tabular}} 
        & Valid JSON & 99.00\% & 99.00\%\\
        & Template Mem. & 99.82\% & 100.0\%\\
        & BB Selection & 99.47\% & 99.81\%\\ 
        \cmidrule(lr){2-4}
        & Valid SMILES & 95.23\% & 97.33\%\\ 
        & Matched Reactants & 70.93\% & 84.02\%\\ 
        & Good Products & 87.02\% & 87.65\%\\ 
        \bottomrule
    \end{tabular}
    \caption{\textbf{Benchmarks of SynLlama inferences using SynLlama models trained with two sets of reaction templates.} Here, both models are fine-tuned on Llama-3.2-1B model with 2M reaction data generate using the same set of training building blocks. We select 1000 molecules for each model: training and testing data are generated using their corresponding reaction templates; ChEMBL data is the same set of 1000 molecules as described in the main text. All SynLlama inferences are run at $T = 0.1$ and $TopP = 0.1$.}
    \label{tab:llm-benchmark}
\end{table}

\begin{table}[H]
\small
    \centering
    \renewcommand{\arraystretch}{1.2} 
    \setlength{\tabcolsep}{10pt}      
    \begin{tabular}{@{} lccccc @{}}
        \toprule
        \textbf{Task} & \textbf{Sampling Method} & \textbf{$K$} & \textbf{$N_{Syn}$} \\ 
        \midrule
        LLM Benchmark & Frozen Only & 5 & 25 \\
        Synthesis Planning & Greedy Sampling & 5 & 25 \\
        Synthesizable Analog & High Only & 10 & 50 \\
        Hit Expansion & High Only & 20 & 100\\
        \bottomrule
    \end{tabular}
    \caption{\textbf{Hyperparameters used for each task.} Under each task name we include the sampling method used for SynLlama inferences as defined in Additional Methodology Details. $K$ represents the number of most similar SMILES string to take during the reconstruction algorithm. $N_{Syn}$ represents the maximum number of synthesis routes to be tracked for each single SynLlama inference during the reconstruction algorithm.}
    \label{tab:example}
\end{table}

\begin{table}[H]
\small
    \centering
    \renewcommand{\arraystretch}{1.05}  
    \setlength{\tabcolsep}{6pt}        
    \begin{tabular}{@{} ll ccc c @{}}
        \toprule
        \multirow{2.5}{*}{Dataset} & \multirow{2.5}{*}{Method} & \multicolumn{3}{c}{\# of Recon. Mol.} & \multirow{2.5}{*}{Morgan Sim.} \\
        \cmidrule(lr){3-5}
         & & Enamine BB & New BB & Total \\
        \midrule
        \multirow{3}{*}{\begin{tabular}[c]{@{}c@{}}Enamine\\ Diversity\end{tabular}} 
            & SynLlama &  691 & 232 & 741 & 0.92 \\
            & SynLlama - druglike & 574 & 100 & 595 & 0.86 \\ 
            & SynLlama - forward & 529 & 50 & 546 & 0.85 \\ 
        \midrule
        \multirow{3}{*}{\begin{tabular}[c]{@{}c@{}}ChEMBL\\ Data\end{tabular}}  
            & SynLlama & 197 & 152 & 287 & 0.68 \\
            & SynLlama - druglike & 124 & 107 & 197 & 0.61 \\ 
            & SynLlama - forward & 132 & 58 & 168 & 0.59 \\ 
        \midrule
        \multirow{4}{*}{\begin{tabular}[c]{@{}c@{}}Branching\\ synthesis \end{tabular}} 
            & ChemProjector  & 302 & - & 302 & 0.79 \\ 
            & SynLlama (RXN 1) & 415 & 118 & 465 & 0.87 \\ 
            & Synformer$\dagger$ & 39 & - & 39 & 0.61 \\
            & SynLlama (RXN 2) & 358 & 101 & 408 & 0.84 \\ 
        \bottomrule
    \end{tabular}
    
    \vspace{0.3cm}
 $\dagger$ The released Synformer\cite{gao_generative_2024} model weights were fine-tuned extensively on smaller drug-like compounds, which caused it to fall short on synthesis planning for more complex molecules.
   \caption{\textbf{Reconstruction performances across various reaction data sets and prompt design choices.} SynLlama-druglike refers to fine-tuning on target molecule that falls within the same distribution as ChEMBL. More detailed analysis of the target molecule properties can be found in Figure S1. SynLlama-forward refers to fine-tune on prompt-response pairs structured as forward synthesis rather than retrosynthesis. SynLlama-branching refers to reconstructions based on tree-like synthesis with testing molecules generated similarly to the training data using testing building blocks and corresponding two sets of reaction templates. We compare to ChemProjector\cite{luo_projecting_2024} (RXN 1) and Synformer\cite{gao_generative_2024} (RXN 2).}
    \label{tab:design-choice}
\end{table}

\begin{table}[H]
\small
    \centering
    \renewcommand{\arraystretch}{1.2}
    \setlength{\tabcolsep}{6pt}
    \begin{tabular}{lcccc}
        \toprule
        \multirow{2.5}{*}{Dataset} & \multirow{2.5}{*}{\% of BB in Enamine} & \multicolumn{3}{c}{\# of Raw Reconstructed Mol.} \\
        \cmidrule(lr){3-5}
        & & Enamine BBs & New BBs & Total \\
        \midrule
        Testing  & 75.85\% & 506 & 125 & 563 \\
        Enamine & 73.51\% & 510 & 100 & 557 \\
        ChEMBL  & 48.07\% & 161 & 95  & 221  \\
        \bottomrule
    \end{tabular}
    \caption{Comparison of Enamine BB presence and reconstruction with purchasable BBs across datasets at greedy temperature and top-p combo when using 91 RXN templates (RXN 1).}
    \label{tab:bb_comparison}
\end{table}

\begin{table}[H]
\small
    \centering
    \renewcommand{\arraystretch}{1.2}
    \setlength{\tabcolsep}{6pt}
    \begin{tabular}{lcccc}
        \toprule
        \multirow{2.5}{*}{Dataset} & \multirow{2.5}{*}{\% of BB in Enamine} & \multicolumn{3}{c}{\# of Raw Reconstructed Mol.} \\
        \cmidrule(lr){3-5}
        & & Enamine BBs & New BBs & Total \\
        \midrule
        Testing  & 76.61\% & 465 & 114 & 520 \\
        Enamine & 68.34\% & 647 & 232 & 711 \\
        ChEMBL  & 48.04\% & 179 & 152  & 280  \\
        \bottomrule
    \end{tabular}
    \caption{Comparison of Enamine BB presence and reconstruction with purchasable BBs across datasets at greedy temperature and top-p combo when using 115 RXN templates (RXN 2).}
    \label{tab:bb_comparison}
\end{table}

\begin{table}[H]
\small
    \centering
    \renewcommand{\arraystretch}{1.05}
    \setlength{\tabcolsep}{6pt}

    \begin{tabular}{@{} ll ccc @{}}
        \toprule
        Dataset & Method & Morgan & Scaffold & Gobbi \\
        \midrule
        \midrule
        \multirow{5.5}{*}{\begin{tabular}[c]{@{}c@{}}Enamine\\ Diversity\end{tabular}}  
            & SynNet  & 0.57 & 0.57 & 0.52\\
            & ChemProjector & 0.82 & 0.85 & 0.83 \\ 
            & Synformer & 0.91 & 0.92 & 0.89 \\ 
        \cmidrule(lr){2-5}
            & SynLlama (RXN 1) & 0.87 & 0.88 & 0.85 \\ 
            & SynLlama (RXN 2) & 0.92 & 0.94 & 0.92 \\
        \midrule
        \midrule
        \multirow{5.5}{*}{\begin{tabular}[c]{@{}c@{}}ChEMBL\\ Data\end{tabular}}  
            & SynNet   & 0.43 & 0.20 & 0.27 \\
            & ChemProjector & 0.60 & 0.59 & 0.56 \\ 
            & Synformer & 0.67 & 0.72 & 0.72 \\ 
        \cmidrule(lr){2-5}
            & SynLlama (RXN 1) & 0.66 & 0.67 & 0.63 \\
            & SynLlama (RXN 2) & 0.68 & 0.69 & 0.66 \\
        \bottomrule
    \end{tabular}
    \vspace{0.3cm}
    \caption{\textbf{Similarity metric comparisons over all successful reconstructions of target and analog molecules.} Similarity metrics using Morgan, Scaffold, and Gobbi similarity scores. SynNet\cite{gao_amortized_2022} and Chemprojector\cite{luo_projecting_2024} are trained using RXN 1, and Synformer\cite{gao_generative_2024} is trained using RXN 2. Scores are computed over successful reconstructions of target and analog molecules from Table 2.}
    \label{tab:model-similarity}
\end{table}

\begin{table}[H]
\small
    \centering
    \renewcommand{\arraystretch}{1.05}  
    \setlength{\tabcolsep}{6pt}        

    \begin{tabular}{@{} ll ccc @{}}
        \toprule
        \multirow{2.5}{*}{Dataset} & \multirow{2.5}{*}{Method}  
        & \multicolumn{3}{c}{Similarity} \\  
        \cmidrule(lr){3-5}  
         &  & Morgan & Scaffold & Gobbi \\  
        \midrule
        \midrule
        \multirow{5.5}{*}{\begin{tabular}[c]{@{}c@{}}Enamine\\ Data\end{tabular}}  
            & SynNet  & 0.51 & 0.51 & 0.45  \\
            & ChemProjector & 0.67 & 0.72 & 0.69 \\ 
            & Synformer & 0.74 & 0.76 & 0.69  \\ 
        \cmidrule(lr){2-5}
            & SynLlama(RXN 1) & 0.69 & 0.72 & 0.65 \\ 
            & SynLlama(RXN 2) & 0.69 & 0.75 & 0.70 \\ 
        \midrule
        \midrule
        \multirow{5.5}{*}{\begin{tabular}[c]{@{}c@{}}ChEMBL\\ Data\end{tabular}}  
            & SynNet   & 0.39 & 0.38 & 0.22 \\
            & ChemProjector & 0.54 & 0.52 & 0.49 \\ 
            & Synformer & 0.59 & 0.65 & 0.65  \\ 
        \cmidrule(lr){2-5}
            & SynLlama(RXN 1) & 0.56 & 0.57 & 0.51 \\
            & SynLlama(RXN 2) & 0.54 & 0.56 & 0.52 \\ 
        \bottomrule
    \end{tabular} \\
    \vspace{0.3cm}

    \caption{\textbf{Similarity metric comparisons over analog molecules when target molecules could not be constructed.} SynNet\cite{gao_amortized_2022} and Chemprojector\cite{luo_projecting_2024} are trained using RXN 1, and Synformer\cite{gao_generative_2024} is trained using RXN 2. Scores are computed over molecules from Table 2 that could not be fully reconstructed. }
    \label{tab:similarity-comparison}
\end{table}

\clearpage
\section*{Supporting Figures}

\begin{figure}[H]
\centering
\includegraphics[width=0.9\textwidth]{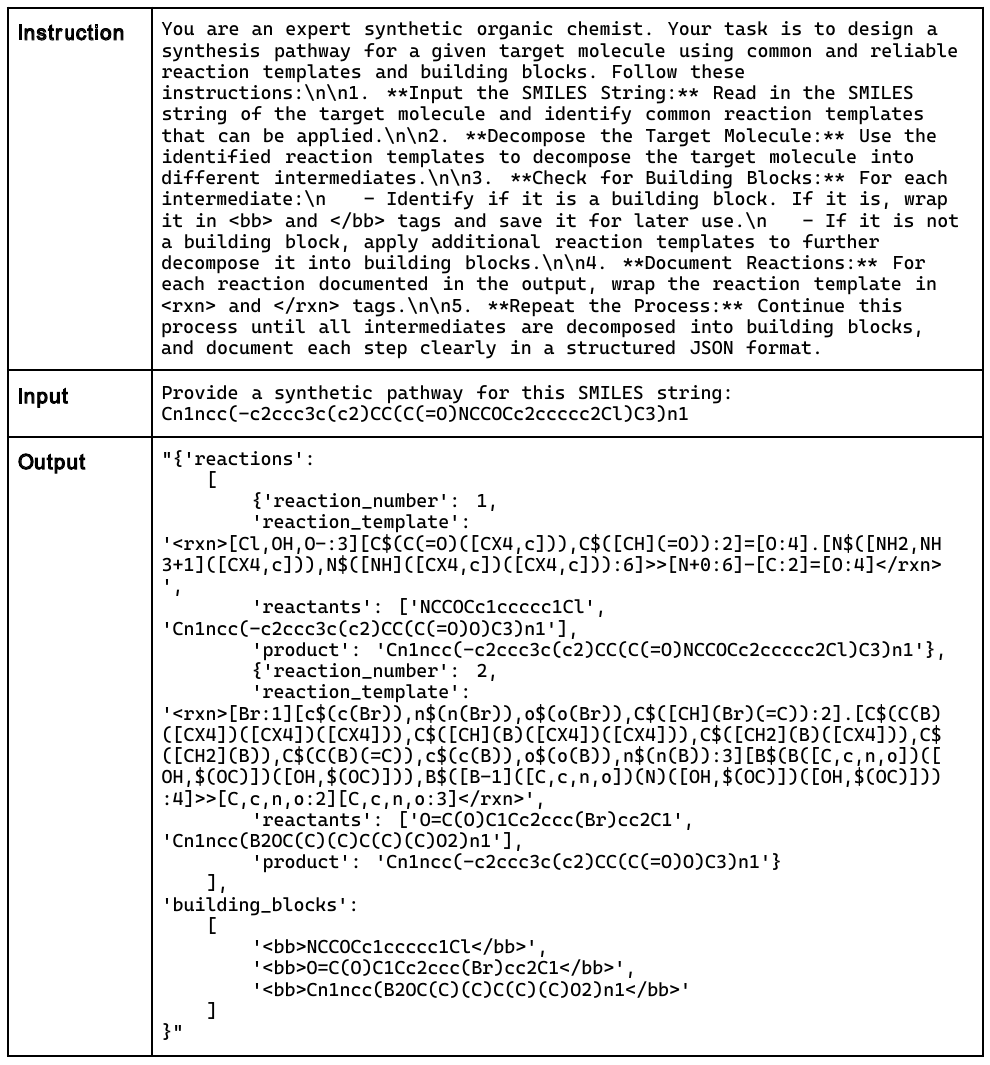}
\caption{\textbf{Instruction, input, and output from the SynLlama model's inference on example SMILES string from Fig.1d.} During data generation, all instructions remain the same, and the input-output pairs are generated within the training synthesizable chemical space. We enforce the JSON format in the output for our post processing algorithms. The output JSON has two parts: reactions and building blocks. In `reactions', a series of reaction steps are generated, where the product of the next reaction serves as the reactant for the previous one. In `building blocks', BBs are selected from the `reaction' section and compiled into a list.}
\label{fig:llm-raw-texts-inference}
\end{figure}

\begin{figure}[H]
\centering
\includegraphics[width=0.75\textwidth]{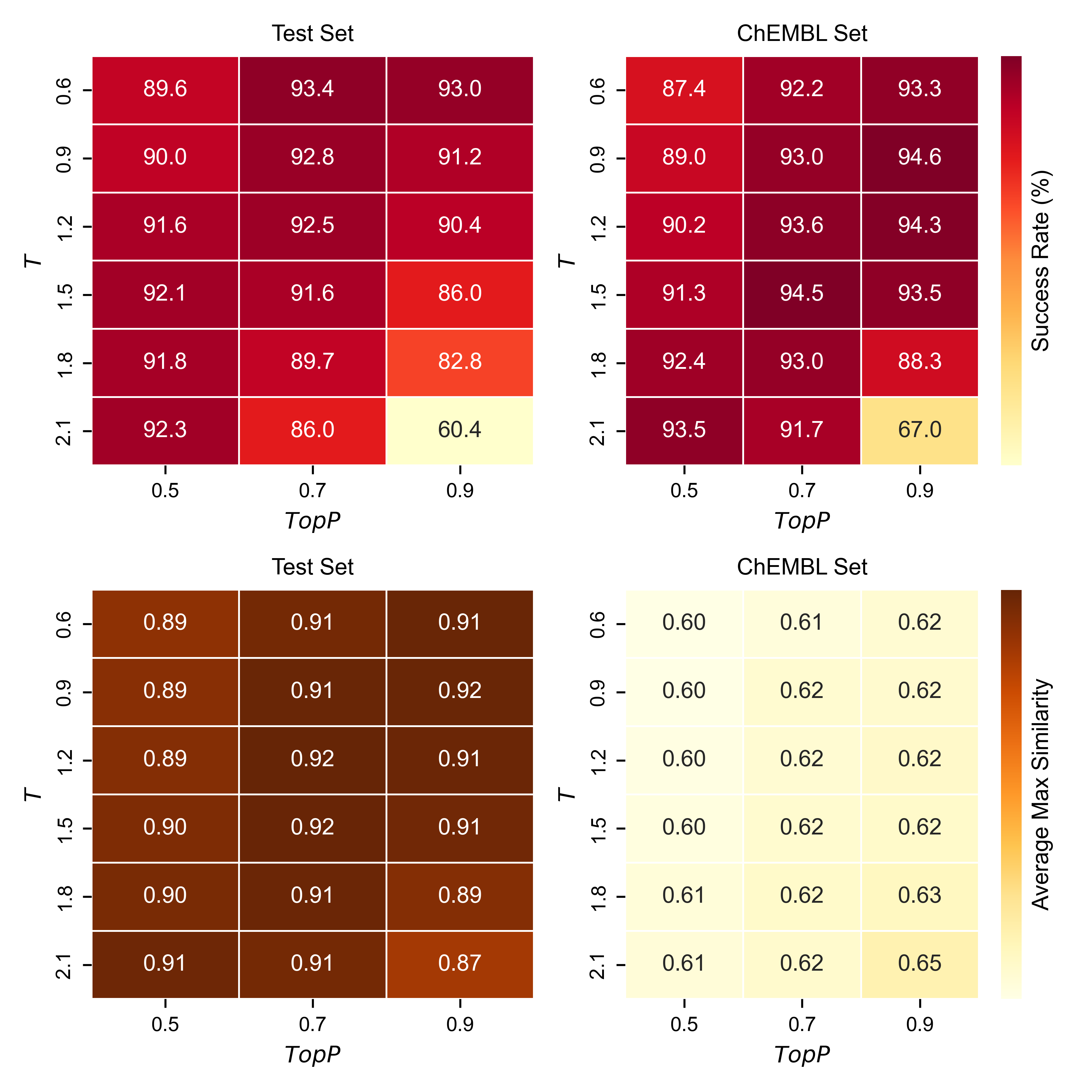}
\includegraphics[width=0.75\textwidth]{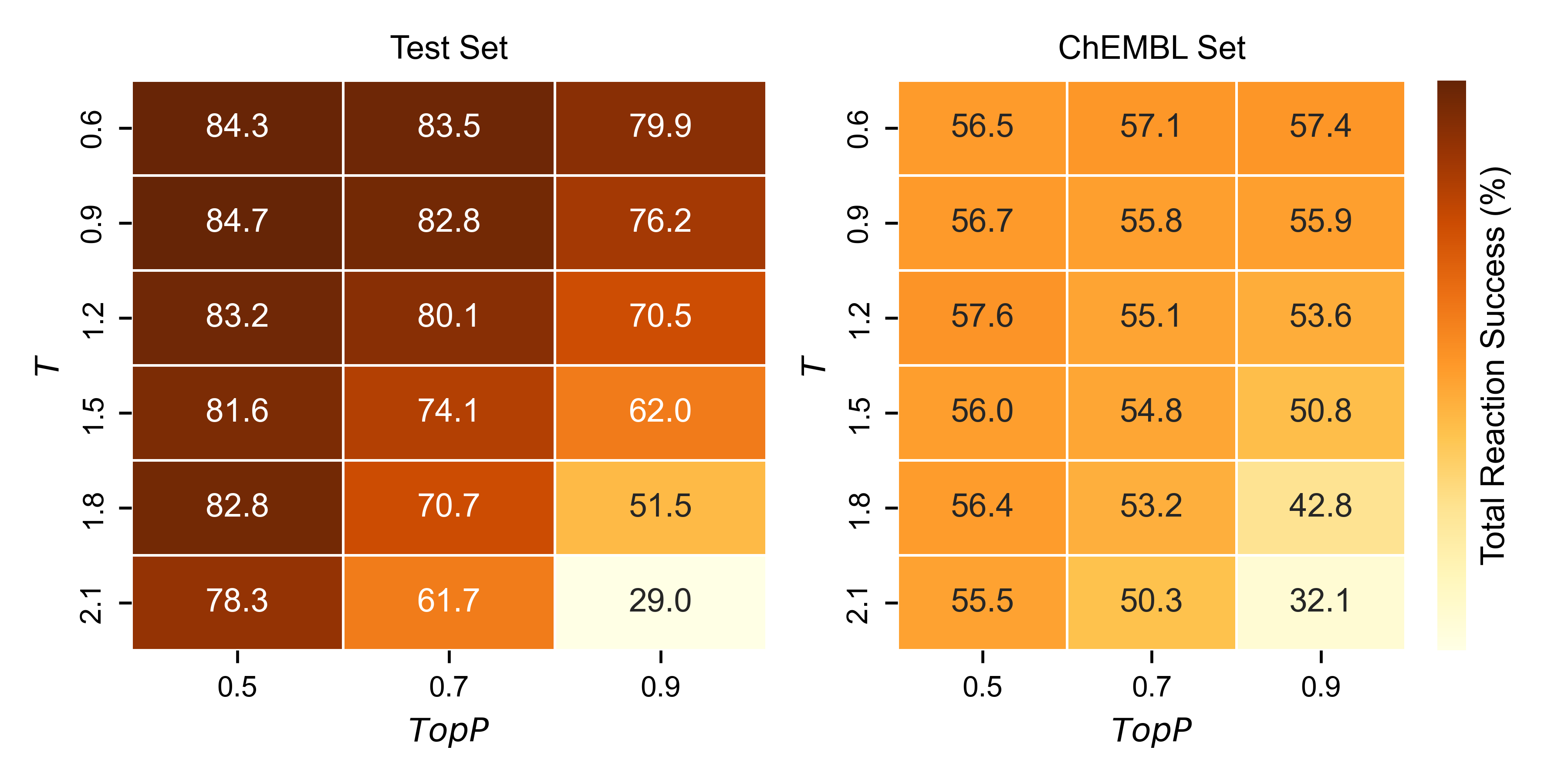}
\caption{\textbf{Reconstruction algorithm and SynLlama raw output benchmarks for SynLlama inferences on the Testing and ChEMBL sets under various temperature and top-p combinations.} The first row represents the success rate of the Enamine reconstruction algorithm based on SynLlama inference outputs. The second row represents the average maximum Tanimoto similarity between the target and analogs generated via the reconstruction algorithm based on 4096-bit Morgan fingerprints. The last row represents the percentage of SynLlama raw outputs that can directly represent a retrosynthetic path for the input molecule without downstream processing with the reconstruction algorithm.}
\label{fig:rxn-success-t-topp}
\end{figure}

\begin{figure}[H]
\centering
\includegraphics[width=0.95\textwidth]{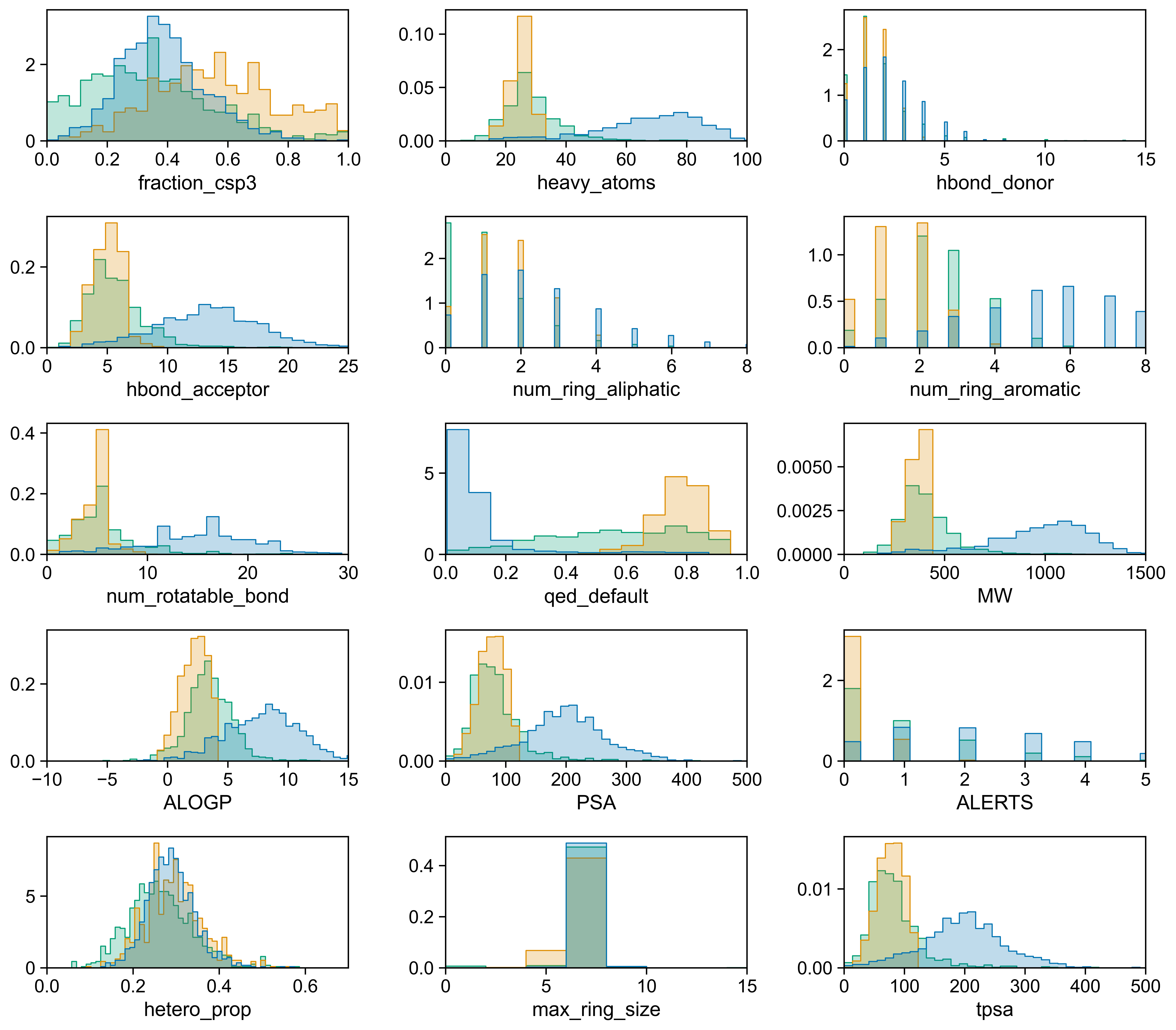}
\caption{\textbf{Drug-related property distributions between product molecules from normal training data (blue), Enamine Diversity Set(orange), and ChEMBL molecules (green).}}
\label{fig:bb-distribution}
\end{figure}

\begin{figure}[H]
\centering
\includegraphics[width=0.95\textwidth]{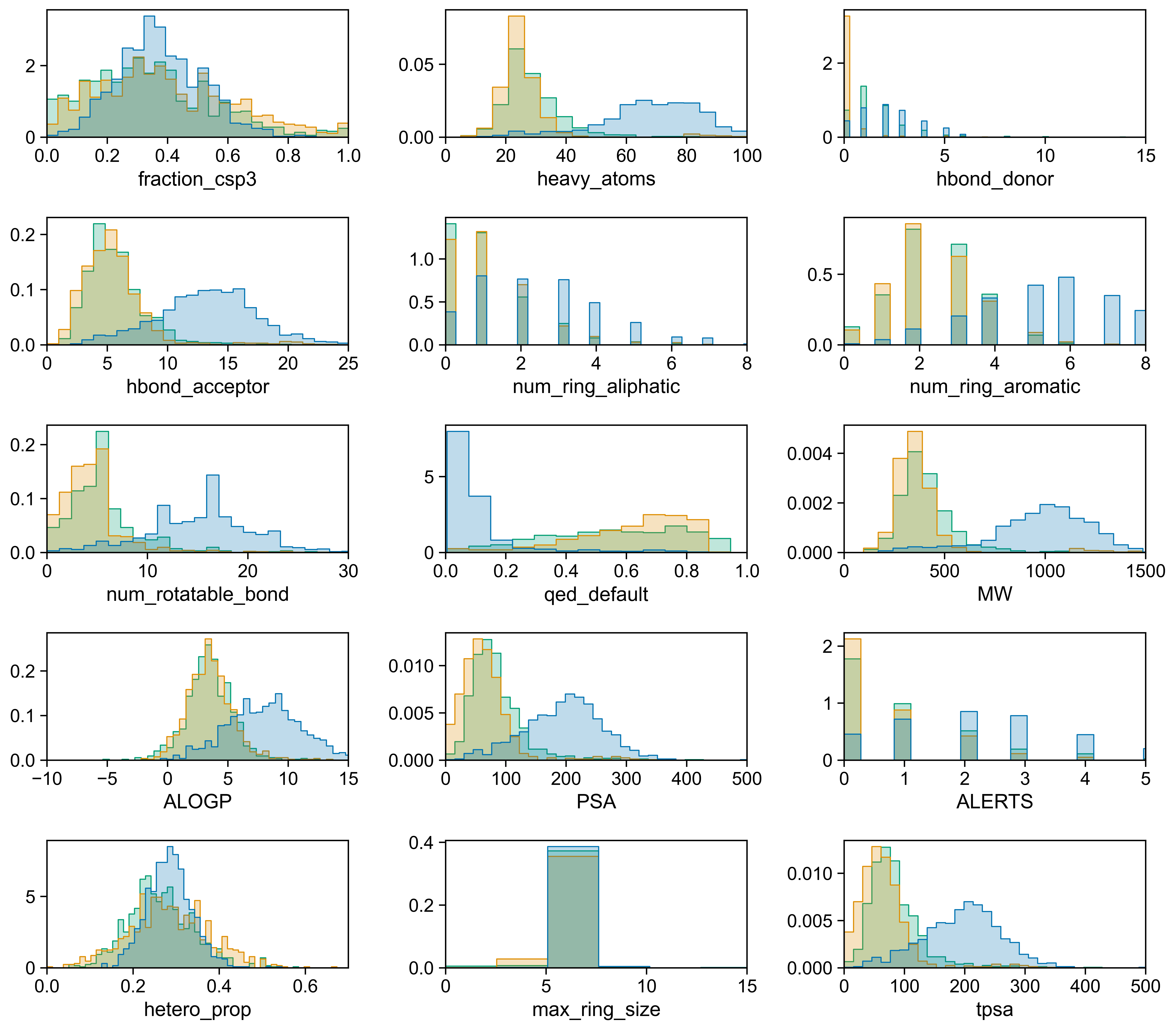}
\caption{\textbf{Drug-related property distributions between product molecules from normal training data (blue), product molecules from training data constrained on druglike properties (orange), and ChEMBL molecules (green).} The generated product molecules under the constraint of druglike properties display similar distribution as ChEMBL molecules. The product molecules from normal training scheme occupies a very different chemical space with more larger molecules.}
\label{fig:bb-distribution}
\end{figure}

\begin{figure}[H]
\centering
\includegraphics[width=0.95\textwidth]{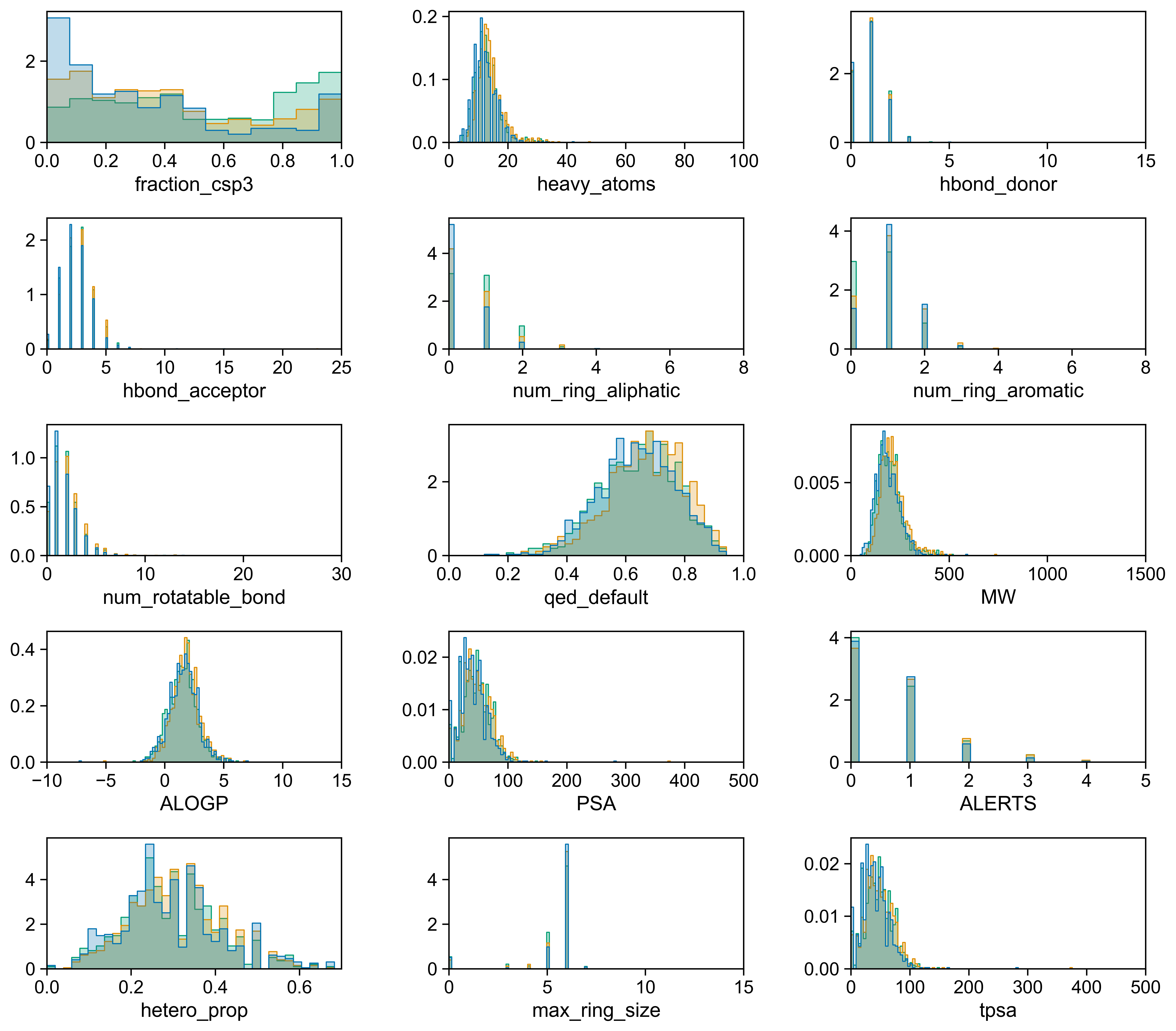}
\caption{\textbf{Drug-related property distributions between training building blocks (blue), testing building blocks (orange), and Molport building blocks (green).} The three building blocks show very similar distribution in all categories.}
\label{fig:bb-distribution}
\end{figure}

\begin{figure}[H]
\centering
\includegraphics[width=0.85\textwidth]{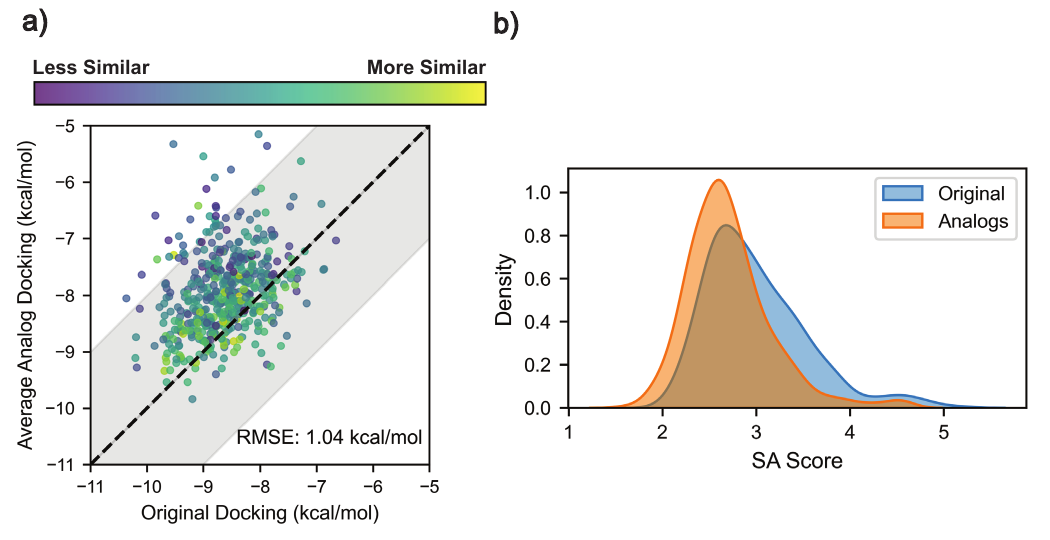}
\vspace{-3mm}
\caption{\textbf{Docking score and SA score distribution between 500 iMiner-generated molecules and proposed analogs from SynLlama model trained on RXN 1.} (a) Correlation plot comparing docking scores of 500 iMiner-generated molecules and the average docking scores of ten most similar analogs for each iMiner-generated molecule. Each data point is color-coded by the average Morgan fingerprint similarity computed between the iMiner target molecules and their corresponding analogs. The shaded area is the energy uncertainty range of $\pm2 kcal/mol$, which is typical for AutoDock Vina scores\cite{Trott2010}. (b) SA score distribution of iMiner molecules and SynLlama-proposed analogs.}
\label{fig:iminer-analog-91}
\end{figure}

\vspace{-8mm}
\begin{figure}[H]
\centering
\includegraphics[width=0.7\textwidth]{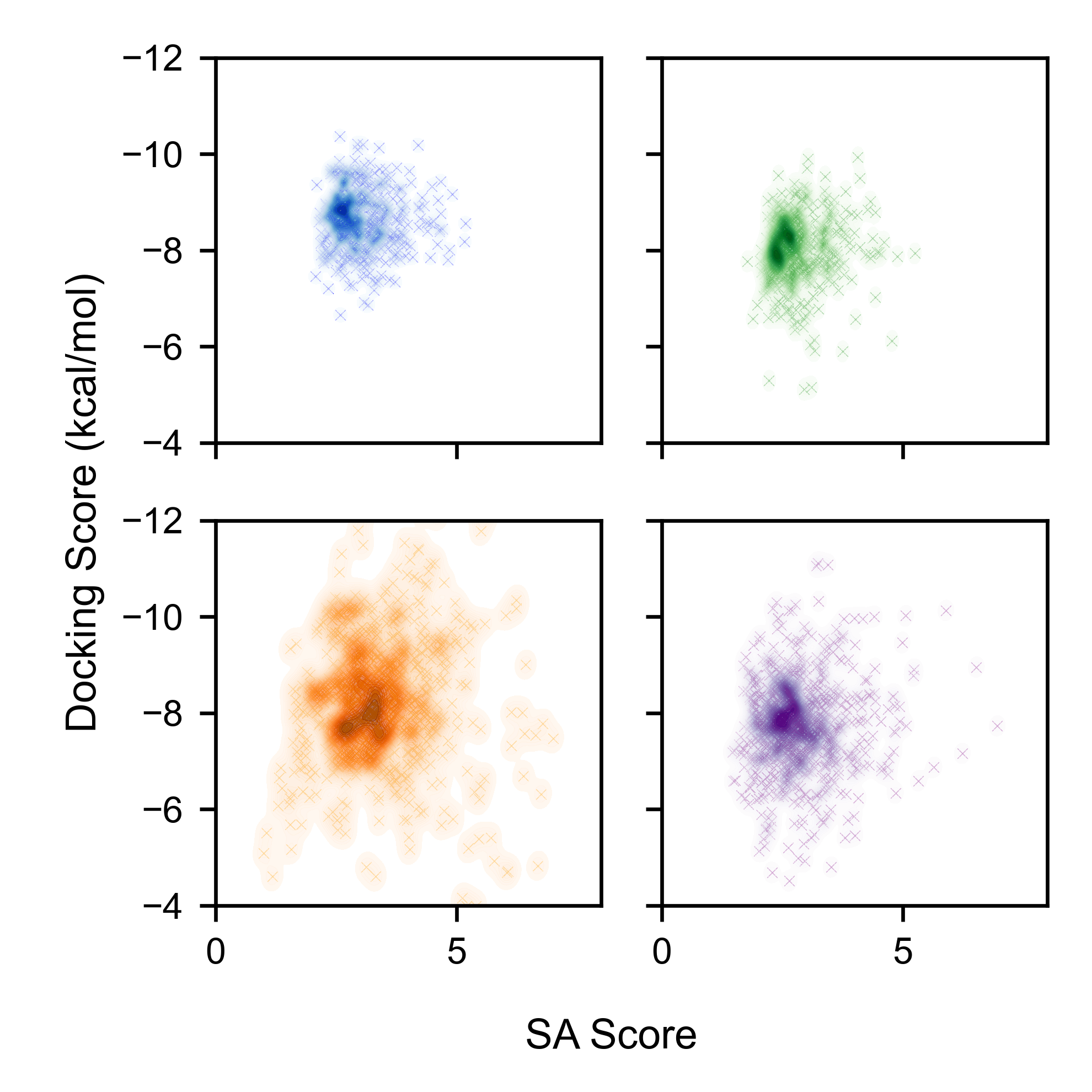}
\vspace{-6mm}
\caption{\textbf{Kernel density estimations of docking scores and SA score for target molecules and SynLlama-generated analogs.} Blue: iMiner targets. Green: iMiner analogs. Orange: Pocket2Mol targets. Purple: Pocket2Mol analogs.}
\label{fig:analog-kde}
\end{figure}

\vspace{-5mm}
\begin{figure}[H]
\centering
\includegraphics[width=0.8\textwidth]{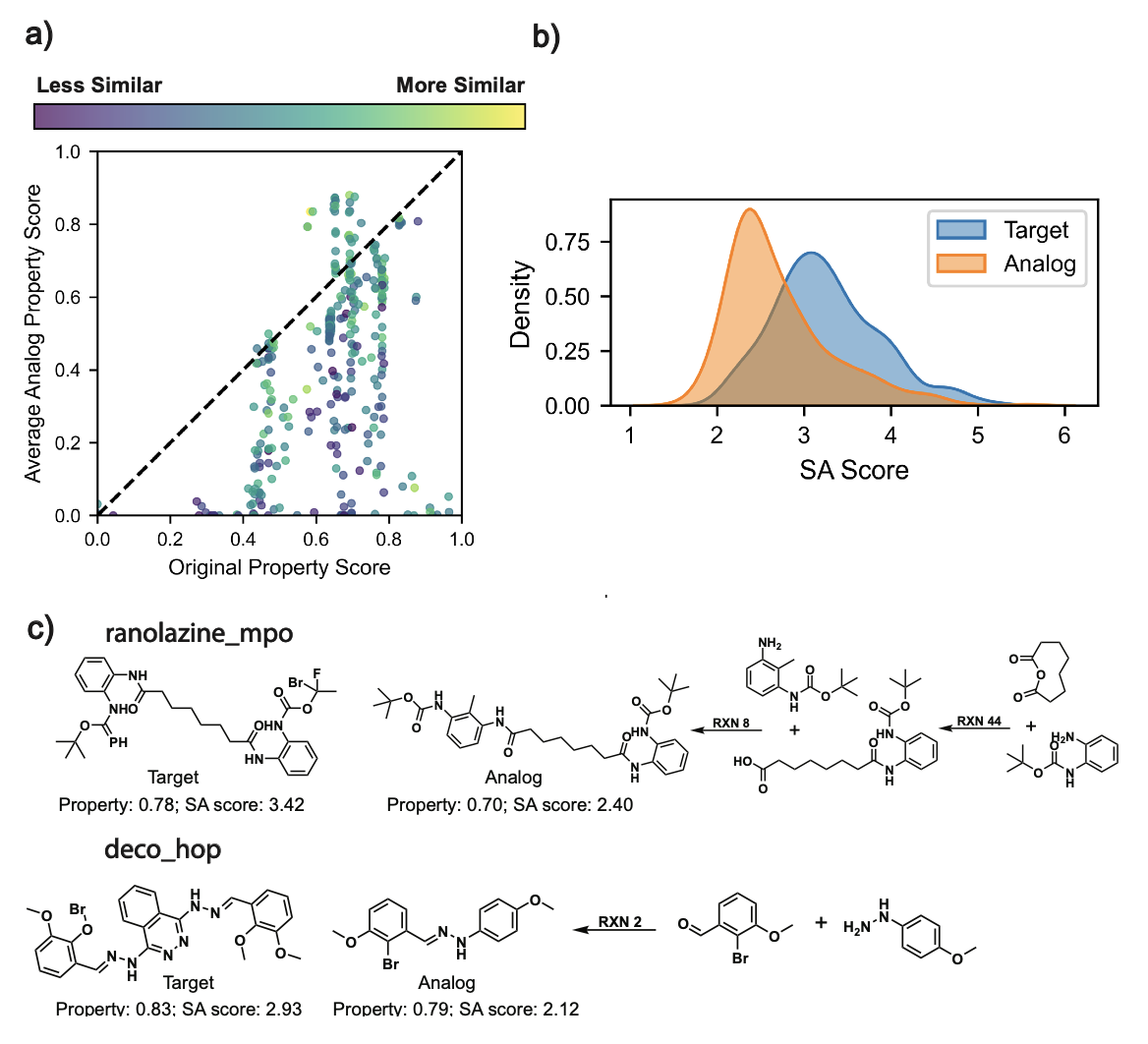}
\vspace{-3mm}
\caption{\textbf{Oracle score and SA score distribution between 500 ASKCOS unsynthesizable molecules and proposed analogs from SynLlama model trained on RXN 2.} (a) Correlation plot comparing property scores of 500 ASKCOS unsynthesizable molecules and the average docking scores of ten most similar analogs for each molecule. Each data point is color-coded by the average Morgan fingerprint similarity computed between the ASKCOS unsynthesizable molecules and their corresponding analogs. (b) SA score distribution of ASKCOS unsynthesizable molecules and SynLlama-proposed analogs. (c) Property and SA scores for example target-analog pairs along with the predicted analog synthetic pathways for two optimization targets: Ranolazine MPO and Deco Hop.}
\label{fig:askcos-analog}
\end{figure}

\begin{figure}[H]
\centering
\includegraphics[width=0.9\textwidth]{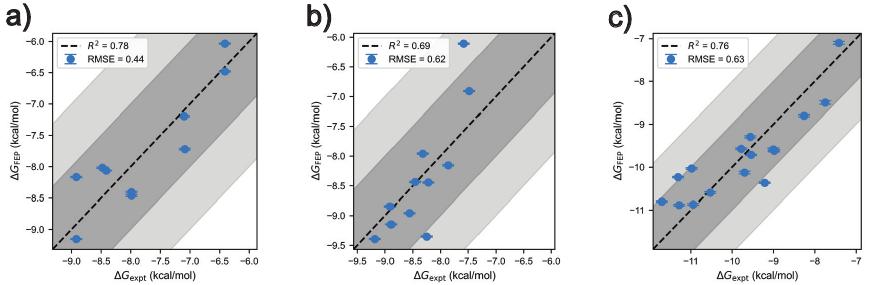}
\vspace{-3mm}
\caption{\textbf{FEP benchmarking on all three protein systems.} Correlation plots between $\Delta G$ extracted experimental IC50 values and $\Delta G$ calculated from FEP for (a) SARS-CoV-2 Mpro\cite{kneller_hit_expansion_7ltj}, (b) Thrombin\cite{thrombin}, and (c) TYK2\cite{tyk2, tyk2_2}. The correlations across all three systems have RMSE $<$ 1 kcal/mol, indicating the reliability of FEP calculations.}
\label{fig:rxn-success-t-topp}
\end{figure}

\bibliography{reference}
\bibliographystyle{naturemag}